%% file: neurips_2025.tex
\documentclass{article}

\usepackage[final]{neurips_2025}

\usepackage[utf8]{inputenc} % allow utf-8 input
\usepackage[T1]{fontenc}    % use 8-bit T1 fonts
\usepackage{hyperref}       % hyperlinks
\usepackage{url}            % simple URL typesetting
\usepackage{booktabs}       % professional-quality tables
\usepackage{amsfonts}       % blackboard math symbols
\usepackage{nicefrac}       % compact symbols for 1/2, etc.
\usepackage{microtype}      % microtypography
\usepackage{xcolor}         % colors
\usepackage{array,multirow,graphicx}
\usepackage{float}
\usepackage{amsmath}
\usepackage{mathtools}
\usepackage{bbm}
\usepackage{hhline}
\usepackage{boldline}
\usepackage{tabularx}
\usepackage{float}
\usepackage{comment}
\usepackage{color}
\usepackage{amssymb}
\usepackage{booktabs}
\usepackage{multicol}
\usepackage{microtype}      % microtypography
\usepackage{xcolor}         % colors
\usepackage{algorithm}
\usepackage{algpseudocode}
\usepackage{pifont}% http://ctan.org/pkg/pifont
\usepackage{soul} %\st
\usepackage{subcaption}
\usepackage{wrapfig}
\usepackage{lipsum}
\usepackage{booktabs}
\usepackage{colortbl}
\usepackage{nicematrix}
\usepackage{kotex}
\usepackage{placeins}
\usepackage{pifont}

\title{Mitigating Semantic Collapse\\in Partially Relevant Video Retrieval}

\author{%
WonJun Moon$^\dagger$, MinSeok Jung$^\dagger$, Gilhan Park, Tae-Young Kim,\\ \textbf{Cheol-Ho Cho, Woojin Jun, Jae-Pil Heo}\thanks{Corresponding author} \\
  Sungkyunkwan University\\
  \texttt{\{wjun0830, alstjr88, a01152a, jackdawson,} \\
  \texttt{gersys, junwoojin, jaepilheo\}@skku.edu} \\
}

\begin{document}

\maketitle
\begingroup
\renewcommand\thefootnote{\dag}
\footnotetext{Equal contribution.}
\endgroup
\input{sec/0_abstract}    
\input{sec/1_intro}

\input{sec/2_related_work}

\input{sec/3_methods}
\input{sec/4_experiments}
\input{sec/5_conclusion}

\section*{Acknowledgements}
This work was supported in part by MSIT/IITP (No. RS-2022-II220680, RS-2020-II201821, RS-2019-II190421, RS-2024-00459618, RS-2024-00360227, RS-2024-00437633, RS-2024-00437102, RS-2025-25442569), MSIT/NRF (No. RS-2024-00357729), and KNPA/KIPoT (No. RS-2025-25393280).

{
    \small
    \bibliographystyle{abbrv}
    \bibliography{reference}
}
\input{checklist}
\newpage
\input{main_appendix}

% {
%     \small
%     \bibliographystyle{abbrv}
%     \bibliography{reference}
% }

% \begin{refsection}
% \input{main_appendix}
% {
%     \small
%     \bibliographystyle{abbrv}
%     \bibliography{appendix-reference} % prints only refs cited in appendix
% }
% \end{refsection}

\end{document}

%% file: sec/0_abstract.tex
\begin{abstract}
Partially Relevant Video Retrieval~(PRVR) seeks videos where only part of the content matches a text query. 
Existing methods treat every annotated text–video pair as a positive and all others as negatives, ignoring the rich semantic variation both within a single video and across different videos. 
Consequently, embeddings of both queries and their corresponding video‐clip segments for distinct events within the same video collapse together, while embeddings of semantically similar queries and segments from different videos are driven apart.
This limits retrieval performance when videos contain multiple, diverse events.
This paper addresses the aforementioned problems, termed as semantic collapse, in both the text and video embedding spaces. 
We first introduce Text Correlation Preservation Learning, which preserves the semantic relationships encoded by the foundation model across text queries.
To address collapse in video embeddings, we propose Cross-Branch Video Alignment~(CBVA), a contrastive alignment method that disentangles hierarchical video representations across temporal scales.
Subsequently, we introduce order-preserving token merging and adaptive CBVA to enhance alignment by producing video segments that are internally coherent yet mutually distinctive.
Extensive experiments on PRVR benchmarks demonstrate that our framework effectively prevents semantic collapse and substantially improves retrieval accuracy.
\end{abstract}

%% file: sec/1_intro.tex
\section{Introduction}
\label{sec.intro}
\begin{figure}
\vspace{-0.2cm}
    \centering
    \includegraphics[width=1\linewidth]{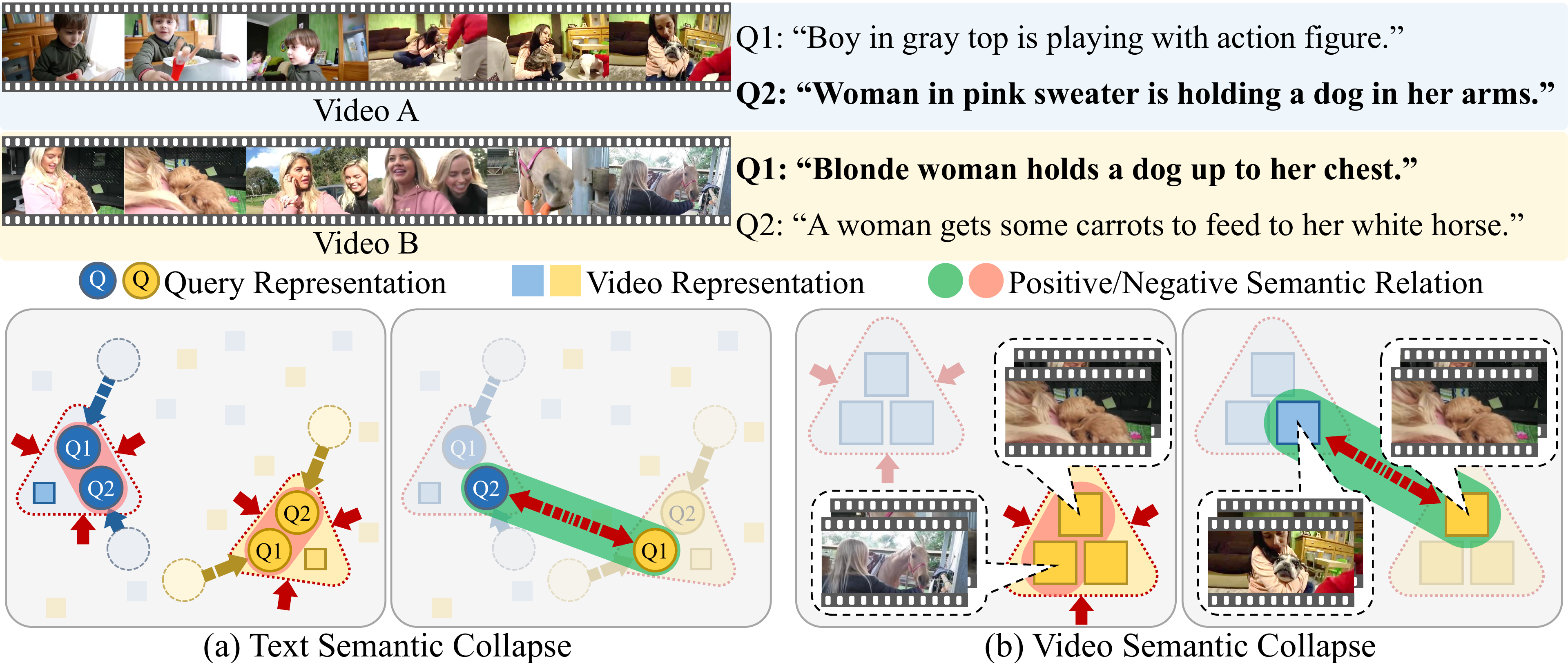}
    \caption{
        Illustration of semantic collapse. 
        (\textbf{Up}) 
        % Untrimmed videos in PRVR encompass multiple moments of different semantics described by multiple text queries. 
        Untrimmed videos in PRVR encompass diverse semantics that can be described by different texts.
        As a result, semantic segments~(both text and video clips) from the same video may convey very different meanings, while segments from different videos can nonetheless be closely related.
        For example, Q2 of Video A and Q1 of Video B both depict “holding a dog”.
        (\textbf{Down}) Since all queries tied to a given video are treated as positives and negative queries drawn from other videos, the model pulls together all text embeddings~(and their corresponding video segments) for that video, regardless of true meaning, and pushes apart semantically similar queries~(and segments) from different videos.
        (a) illustrates that queries of the same video are pulled together regardless of their semantic relationships~(left), while queries with similar context~(holding a dog) are pushed apart~(right).
        (b) shows that video segments also suffer from the same phenomenon.
    % The fundamental challenges of PRVR and the results of our proposed method. 
    % In (a), queries are mapped based on their respective video tokens, where Q2 of Video A and Q1 of Video B are relatively positive, while all other dissimilar queries are relatively negative. However, existing PRVR methods suffer from two key issues: (b) intra-semantic collapse and (c) inter-semantic collapse. In (b), this occurs when queries within the same video are overly similar due to pairwise annotations, even if they have a negative relationship. In (c), this arises when semantically similar queries from different videos are pushed apart, as the model treats them as negatives, degrading retrieval performance. To address these issues, our approach (d) leverages (a) well-structured embedding space of CLIP to preserve semantic relationships while maintaining query distinctiveness.
    }
    \label{fig:motiv_semanticcollapse}
    \vspace{-0.4cm}
\end{figure}
Recently, Partially Relevant Video Retrieval~(PRVR)~\citep{ms-sl_PRVR, gmm_PRVR, gmmv2_PRVR}  has emerged as a significant research challenge in computer vision. 
PRVR shares the same objective as traditional Text-to-Video Retrieval~\cite{Alea_VR, CLIPPING_VR, MMCKD_VR, Pidro_VR,VoP_VR,X-clip_VR}, retrieving the video that best aligns with a given text query. 
However, the key difference lies in PRVR's assumption that target videos may be only partially relevant to the query rather than requiring a perfect semantic match.
The primary challenge in PRVR lies in learning from text-video pairwise annotations. 
A single video is often associated with multiple distinct text queries labeled as positive pairs; however, the semantic relationships among these text queries are not explicitly defined, and fine-grained temporal annotations that indicate their precise alignment within the video are typically unavailable.

As a result, conventional training for retrieval based on the InfoNCE loss~\cite{infonce_Loss, supercon} induces a semantic collapse problem in PRVR. 
Semantic collapse refers to the phenomenon where paired text queries and visual segments are excessively attracted to each other while being indiscriminately repelled from features of other pairs, regardless of their actual semantic similarity.
Fig.~\ref{fig:motiv_semanticcollapse}~(a) illustrates this issue within the text embedding space; text queries associated with the same video tend to cluster together even when they are semantically unrelated, while semantically similar queries are pulled apart when they are paired with different videos.
In addition, the same phenomenon occurs in video embeddings; video segments drawn from the same video collapse together regardless of their true semantic differences, as shown in Fig.~\ref{fig:motiv_semanticcollapse}~(b).
This is because the training guidance is provided by video ID, not by their individual semantic content.
In short, every segment in a video shares the identical set of paired text queries as positives.

Previous works, e.g., GMMFormer~\cite{gmm_PRVR} and GMMFormer-v2~\cite{gmmv2_PRVR}, have attempted to address the semantic collapse within text embeddings.
Specifically, these methods explicitly reduce the similarity between text queries paired with the same video.
However, the semantic relationships between text queries are often overlooked, and the issue of semantic collapse within video embeddings remains underexplored, leading to sub-optimal performance.

In this paper, we aim to mitigate the semantic collapse in both text and video embeddings for PRVR.
First, we introduce Text Correlation Preservation Learning~(TCPL), which leverages CLIP~\cite{clip_ID}, a vision-language foundation model with a well-structured semantic space.
By distilling the semantic relationships encoded in CLIP, TCPL effectively regularizes the semantic collapse within text embeddings.
While TCPL leverages CLIP’s rich text‐semantic structure to regularize collapse in the textual embedding space, we point out that the same approach cannot be directly applied to video embeddings. 
This is because CLIP’s pretraining operates on static images, thereby lacking the capacity to model temporal dynamics~\cite{li20254d}.

To this end, we introduce Cross-Branch Video Alignment~(CBVA), a dedicated objective to preserve context diversity in the video modality.
CBVA utilizes a dual-branch architecture commonly adopted in PRVR to encode hierarchical video representations and employs a contrastive objective to differentiate distinct events within a video. 
Concretely, frame- and clip-level embeddings from the same timestamp are encouraged to align closely, while those from different timestamps are driven apart.
Then, we further leverage the token merging strategy in two ways to enhance video-adaptivity within CBVA; (1) order-preserving token merging is introduced for semantically consistent video clip aggregation, and (2) bipartite token merging~\cite{tome} is leveraged to organize representative contexts within each video.
By encoding clips in a context-aware manner, we encourage videos to be represented in line with their true semantic content.
Consequently, with TCPL and CBVA combined, our method achieves state-of-the-art performances in all tested benchmarks.

In summary, our contributions are (1) We propose Text Correlation Preservation Learning, which leverages the semantic relationships within the foundation model to address semantic collapse within text embeddings, (2) We propose Cross-Branch Video Alignment to mitigate the semantic collapse in video modality by distinguishing distinct events within a video, (3) We leverage token merging strategies to encourage the precise video alignment, and (4) Our method achieves superior performances across all datasets in PRVR.

%% file: sec/2_related_work.tex
\section{Related Work}
\textbf{Partially Relevant Video Retrieval.}
PRVR aims to retrieve untrimmed videos that are partially relevant to a given query~\cite{ms-sl_PRVR, pean, jun2025bridging, mgakd}.
MS-SL~\cite{ms-sl_PRVR} addresses this challenge by proposing a dual encoding strategy that explicitly separates features for frame and clip segments, capturing different temporal scales within untrimmed videos.
Subsequently, DL-DKD~\cite{dldkd_PRVR} leverages CLIP~\cite{clip_ID} to enhance PRVR performance by distilling text–frame similarity.
GMMFormer~\cite{gmm_PRVR} introduces a Gaussian Mixture Model–based Transformer that enables efficient retrieval with a reduced set of video features. 
It also identifies semantic collapse as a key challenge and proposes a query-diverse loss to enforce separation among multiple text queries linked to the same video. 
Building on this, GMMFormer v2~\cite{gmmv2_PRVR} further addresses semantic collapse by explicitly controlling the degree of semantic separation between queries associated with the same video.
% Seeking an efficient retrieval framework, GMMFormer~\cite{gmm_PRVR} introduces a Gaussian Mixture Model-based Transformer, enabling retrieval with fewer video features; it also highlights the issue of semantic collapse and proposes a query-diverse loss to ensure multiple text queries associated with the same video to be separated.
% GMMFormer v2~\cite{gmmv2_PRVR} further addresses semantic collapse by manipulating the degree of semantic separation between queries paired with the same video..
% GMMFormer v2~\cite{gmmv2_PRVR} further addresses semantic collapse by separating semantics based on the similarity among encoded text features and employing an optimal matching loss so that each query aligns with distinct frames.
Unlike these methods that only enforce separation among a small set of queries, our approach aims to leverage their true semantic relationships and additionally mitigates semantic collapse in the video embedding space.
% Unlike previous approaches that simply reduce semantic similarity to address the semantic collapse, we propose a learning objective that considers the semantic relationships between texts.
% However, we point out that the potential positive pairs between queries, whether within the same video or across different videos, are yet to be considered.
% However, these methods still concentrate on diverging text representations matched with the same video, leaving semantic collapse unresolved.
% To address this semantic collapse, we propose to transfer the well-structured semantic space from the foundational model. 
%-------------------------------------------------------------------------

\textbf{Knowledge Distillation.}
The aim of knowledge distillation is to train a student model with fewer parameters to achieve performance comparable to a larger teacher model~\cite{initial_KD}. 
For classification tasks, Kullback-Leibler divergence loss is widely applied to align the student’s output distribution with that of the teacher after the softmax layer, allowing the student model to learn from the teacher’s predictions.
Subsequently, transferring knowledge at the intermediate feature level has been the next stream~\cite{pay_F_KD,show_F_KD,knowledge_F_KD}. 
However, as they fail to effectively capture the relationships between individual features, Relational Knowledge Distillation (RKD)~\cite{relational_R_KD, instance_R_KD, contrastive_R_KD} was proposed to distill the relationships within the semantic space of the teacher model to that of the student.
In PRVR, the problem of semantic collapse occurs due to the lack of consideration for relationships among queries paired with the same video, as well as across queries from different videos.
Therefore, we leverage RKD to transfer structured semantic relationships within the foundational model to typical PRVR network designs~\cite{ms-sl_PRVR,gmm_PRVR,gmmv2_PRVR} that often suffer from semantic collapse.

% \paragraph{Hierarchical Alignment}

\textbf{Token Merging.}
Token merging~\cite{tome, tomediffusion, algm} has been proposed to improve the efficiency of Transformer~\cite{transformer} by reducing token redundancy. 
A representative method, ToMe~\cite{tome}, uses bipartite matching on token similarities to merge spatial tokens in the vision transformer. 
Recently, token merging strategies have been extended to the video domain. 
For example, LearnableVTM~\cite{longvideo} learns per-patch saliency scores and applies for merging across long videos.
TempMe~\cite{tempme} sequentially merges tokens within progressively larger fixed-window clips, addressing both spatial and temporal redundancy for retrieval.
In contrast, our work applies token merging for two purposes: we merge semantically-coherent adjacent video frames to assemble coherent contexts in each video clip, and leverage token merging to determine the representative context within each video.
These facilitate precise alignment between hierarchical video representations.
 % Inspired by these developments, we adopt a token merging approach to improve clip aggregation strategies of PRVR and mitigate semantic collapse issues inherent to the video modality.
% handles long-term video inputs by dynamically merging tokens based on diverse criteria such as region, motion, and saliency.

%Token merge는 vision transformer (vit)의 efficiency를 개선하기 위한 방법이다 [Tome, tome diffusion faster, vidtome]. 대표적인 방법인 [Tome]는 token간의 similarity가 높은 token을 bipartile matching 을 통해 redundant한 token을 combine하여 vit의 efficency를 높였다. 이러한 token merge 방법은 최근 image 관련 분야에서 video field로 확장되었는데, [tempme]는 progressively하게 negihboring clip token merge를 통해 temporal redundancy를 해결하였고, [longvideo]는 region, sailency 등의 다양한 기준을 바탕으로  dynamically한 token merge 방법을 통해 long-term video input을 효고적으로 처리하였다. 우리는 이전 PRVR 방법들이 사용하던 clip aggregation strategy를 보완하고 video modality의 semantic collapse를 해결하기 위해 token merge 방법을 사용한다.  

% WJ: 흐름 괜찮은거같고 tome 관련된거 몇개만 더 첫문장에 cite 추가해주면 좋을듯. 설명추가는 안해도됨 일단.

% In our work, to address the semantic collapse of text and video features arising from the PRVR objective, we apply Relational Knowledge Distillation (RKD) from the well-established text and video relationships  of CLIP and [??] to existing collapsed PRVR networks~\cite{ms-sl_PRVR,gmm_PRVR,gmmv2_PRVR}. 

\begin{figure*}
    \centering
    \vspace{-0.cm}
    \includegraphics[width=1.\textwidth]{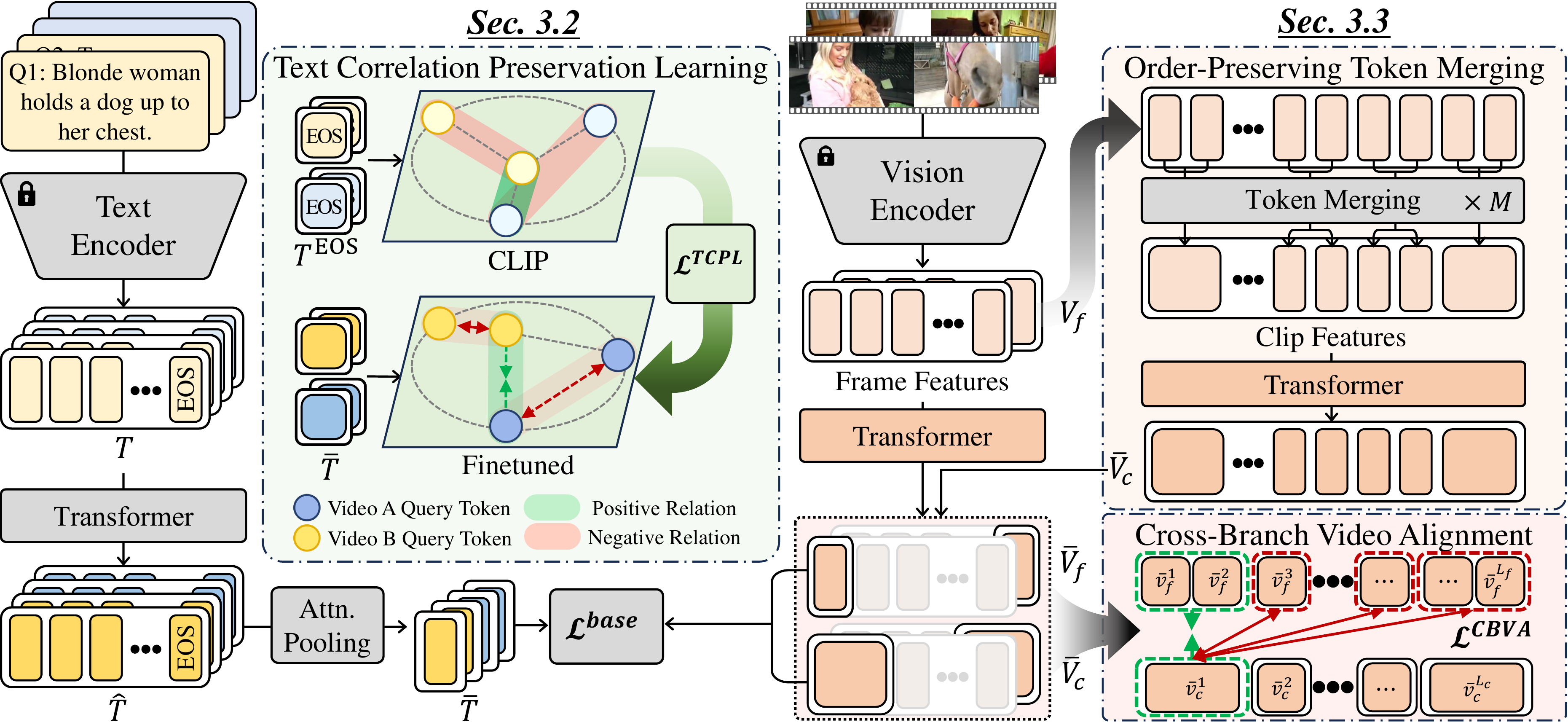}
    \caption{
        Method overview. 
        We extract text and visual tokens with pretrained backbones, which are then processed via transformer layers. 
        Text tokens are aggregated via attention pooling to produce a single query token $\bar{T}$ for each text query.
        Also, following prior works, dual-branch visual tokens are encoded~(both frame- and clip-level), producing a sequence $\bar{V}$ of video tokens for each level.
        A baseline retrieval loss $\mathcal{L}^{\text{base}}$ aligns $\bar{T}$ with the most similar video token at each level.
        To mitigate text-side semantic collapse, Text Correlation Preservation Learning transfers CLIP's query relationships.
        On the other hand, Cross-Branch Video Alignment aligns hierarchical segments by timestamping to mitigate collapse and preserve visual details.
        Furthermore, CBVA is precisely enhanced by constructing coherent clips with Order-Preserving Token Merging and improving adaptivity~(illustrated in Sec.~\ref{Sec.vr}).
        % To mitigate overfitting in text and video encoding, we introduce Text Correlation Preservation Learning for maintaining semantic relationships in text queries and a Self-Relational Transformer to enhance temporal modeling in video representations.
    }
    \label{fig:Overview}
    \vspace{-0.2cm}
\end{figure*}

%% file: sec/3_methods.tex
\section{Method}
\label{Sec.method}

\subsection{Preliminary}
\label{Sec.preliminary}
Our architectural design is illustrated in Fig.~\ref{fig:Overview}. 
Similar to prior works, we employ pretrained encoders to extract tokens, which are processed through trainable layers.

\textbf{Text encoder.}
Given a batch of text inputs, we utilize the pre-trained text encoder to extract text tokens $T \in \mathbb{R}^{B_q \times L_q \times d_{q}}$, where $B_q$, $L_q$ and $d_q$ denote the number of text queries, the number of words per query, and the dimension of query representation, respectively.
% Given a batch of text inputs, we utilize the pre-trained CLIP-L/14~\cite{clip_ID} text encoder to extract text tokens $T \in \mathbb{R}^{B_q \times L_q \times d_{\text{clip}}}$, where $B_q$, $L_q$ and $d_{\text{clip}}$ denote the number of text queries, the number of words per query, and the CLIP representation dimension, respectively.
The sequence of word tokens includes [SOS]~(start of sequence) at the beginning and [EOS]~(end of sequence) at the end, making the total number of tokens $L_q$.
These tokens are forwarded through projection layers and transformer layers to produce text representations $\hat{T} \in \mathbb{R}^{B_q \times L_q \times d}$ for downstream text-video retrieval, where $d$ denotes the projected dimension.
Finally, attention pooling is applied to $\hat{T}$ to derive a single aggregated token $\bar{T} \in \mathbb{R}^{B_q \times d}$ that represents the final representation of the text query.
% For $i$-th text query, attention pooling is expressed as:
% \begin{align}
% \label{eq1}
%     % \bar{T}=\sum_{i=0}^{L_q}\alpha_i \times \hat{t}_i, \alpha= \text{softmax}(\hat{T}{w_{ap}}),\\
%     \bar{T}_i=\sum_{j=0}^{L_q}\alpha_i \times \hat{T}_{i,j}, \quad \alpha_i = \text{softmax}(\hat{T}_i W^T),
% \end{align}
% where $W \in \mathbb{R}^{1 \times d}$ denotes a projection layer and $\alpha \in \mathbb{R}^{ L^q \times d}$ indicates weights for each text token for attention pooling.
% $\hat{T}_{i,j}$ represents the token of the $j$-th word token in $i$-th text query.

\textbf{Video encoder.}
For a batch of $B_v$ videos with $L_f$ frames each, we utilize the pre-trained image or video encoder to extract a visual token~(e.g. [CLS] token from CLIP) for each frame, generating frame tokens $V_f \in \mathbb{R}^{B_v \times L_f \times d_v}$.
Additionally, to represent moments of varying temporal lengths, the frame tokens $V_f$ are aggregated into video clips in the clip branch, to generate clip-level tokens $V_c \in \mathbb{R}^{B_v \times L_c \times d_{v}}$, where $L_c$ denotes the number of clips per video.
Note that our clip construction process is performed with order-preserving token merging, which is discussed in Sec.~\ref{Sec.vr}.
Then, each frame and clip token is encoded independently through the transformer layers to capture contextual relationships.
% Subsequently, temporal modeling among visual tokens is conducted through transformer layers to capture contextual relationships.
Consequently, $\bar{V}_{f} \in \mathbb{R}^{B_v \times L_f \times d}$ and $\bar{V}_{c} \in \mathbb{R}^{B_v \times L_c \times d}$ are produced for final video representations.
% % 이후에 프레임 간 temporal modeling을 수행하기 위해 self-attention layer를 통과한다.
% \begin{equation}
%     {V_i} = [V^i_1, V^i_2, \dots, V^i_{L_i}], \text{where} \ {i} = f, c
% \end{equation}

% 비디오는 다양한 길이의 clip에 대응하기 위해서, 기존의 frame feature와 frame feature를 일정한 길이 k로 mean pooling한 clip feature를 만들어 독립적인 인코딩을 수행한다.
% \begin{equation}
%     \bar{c_i} = \frac{1}{K} \sum_{i=1}^{K} \bar{f}_i, \ \text{where} \ {K} = \frac{L_v}{L_c}
% \end{equation}
% \begin{equation}
%     v_c = [\bar{c}_1, \bar{c}_2, \dots, \bar{c}_{L_c}]
% \end{equation} 
% The two video tokens are encoded independently, but the process is identical.
% 두 feature의 인코딩 과정은 대부분 동일하다.
% Video tokens also forward through the projection layer to align their dimensions with text tokens.
% % video 토큰도 text 토큰과 차원을 맞춰주기 위해 projection layer를 통과한다.
% Subsequently, to perform temporal modeling across video frames or clips, the video tokens forward a self-attention layer.
% % 이후에 프레임 간 temporal modeling을 수행하기 위해 self-attention layer를 통과한다.
% \begin{equation}
%     {V_i} = [V^i_1, V^i_2, \dots, V^i_{L_i}], \text{where} \ {i} = f, c
% \end{equation}
% \begin{equation}
%     V = [\bar{}_1, \bar{f}_2, \dots, \bar{f}_{L_f}]
% \end{equation}
% frame feature는 모델에 따라 frame aggregation module을 통해 하나의 video embedding으로 표현되기도 하며, clip embedding은 다양한 길이 단위로 mean pooling되기도 한다.
\textbf{Training objective.}
To retrieve a video with the given text query, we perform similarity matching between the representations from two modalities.
Specifically, during training, we first select one video token per video that yields the highest similarity to the given text query in both frame and clip branches.
Then, these video tokens~(one from each video representation) are used to conduct retrieval for training using InfoNCE loss~\cite{infonce_Loss, supercon} and triplet ranking loss~\cite{triplet_Loss}.
Accordingly, the final training objective is formulated as follows.
\begin{align}
    \mathcal{L}^\text{base} = \mathcal{L}_c^\text{nce} + \mathcal{L}_c^\text{trip} + \mathcal{L}_f^\text{nce} + \mathcal{L}_f^\text{trip},
\end{align}
where $\mathcal{L}^\text{nce}_*$ and $\mathcal{L}^\text{trip}_*$ indicate the InfoNCE loss and triplet ranking loss, respectively, and $\mathcal{L}_c^*$ and $\mathcal{L}_f^*$ represent the clip-level loss and frame-level loss, respectively.

% For model training and retrieval, given text tokens and video tokens, the similarity score between the two modalities is measured.
% Additionally, the video token with the highest score in a given video is assumed to represent the text-relevant moment.
% % 그리고 한 비디오에서 점수가 가장 높은 video token 하나에 대해 text와 관련된 moment로 가정한다. 
% In other words, the max-pooled video token is extracted.
% % 따라서 해당 moment와 text를 align하기 위해 text token에 대해서 maxpooling된 video token을 추출하여 학습에 활용한다.
% To align text and video embeddings, pairwise labels are used, where positive pairs are assigned based on the labels, and all others are treated as negative pairs. The model is trained using InfoNCE loss \cite{} and Triplet Ranking Loss \cite{}.
% 여기서 text와 video embedding을 align 하기 위해서 pairwise label에 대해 positive pair로 두고 나머지는 negative pair로 활용하여 infoNCE loss[] 와 triplet ranking loss[]를 사용한다.
% Depending on the model, diversity loss [ ] may also be used to mitigate the intra-semantic collapse problem.
% 이후에 모델에 따라 intra semantic collpase 문제를 해결하기 위해 diversity loss[]를 사용하기도 한다.
% Accordingly, the final training objective is formulated as follows.
% % 이에 따라 최종적인 training objective는 다음과 같다.
% \begin{align}
%     \mathcal{L}^\text{retrieval} = \mathcal{L}_c^\text{nce} + \mathcal{L}_c^\text{trip} + \mathcal{L}_f^\text{nce} + \mathcal{L}_f^\text{trip}
% \end{align}

\textbf{Problem definition: semantic collapse.} 
Existing PRVR approaches suffer from semantic collapse which indicates that the general relationships among queries and videos are disrupted.
This phenomenon occurs because pairwise text-video annotations~(which only specify positive relationships) are used for learning PRVR.
Specifically, in PRVR, each video is associated with multiple distinct text queries, which triggers the typical contrastive learning to encourage the queries paired with the same video to cluster together, while text queries paired with different videos are separated as they are attracted to different videos.
In this work, we attempt to alleviate the semantic collapse within the text embedding in Sec.~\ref{Sec.tcpl} and video embedding in Sec.~\ref{Sec.vr}.

\subsection{Semantic Collapse in Text Embeddings: Text Correlation Preservation Learning}
\label{Sec.tcpl}
Previously, GMMFormer~\cite{gmm_PRVR} and GMMFormer-v2~\cite{gmmv2_PRVR} have attempted to address semantic collapse in that they enforced separation between text queries paired with the same video. 
However, we argue that they only partially alleviate the semantic collapse since all text queries paired with the same video are pushed apart without considering their actual semantic relationship.

To mitigate this issue, we propose Text Correlation Preservation Learning~(TCPL), which leverages the well-structured semantic space of CLIP.
Specifically, TCPL explores the semantic relationships between text queries within the CLIP semantic space and distills the relationships toward the retrieval space.
In this work, we measure the relationships with two metrics: Euclidean distance and angular distance. 
These two metrics are defined with the pair $(\mathbf{x}, \mathbf{y})$ and triplet $(\mathbf{x}, \mathbf{y}, \mathbf{z})$, where $\mathbf{x}, \mathbf{y}$, and $\mathbf{z}$ denote text embeddings, respectively, as follows:
\begin{equation}
    f^{\text{e}}(\mathbf{x}, \mathbf{y})=\frac{1}{\mu}\|\mathbf{x} - \mathbf{y}\|_2 \; ; \;
% \end{equation}
% \begin{equation}
    f^{\text{a}}(\mathbf{x}, \mathbf{y}, \mathbf{z})= \left\langle \frac{\mathbf{x} -\mathbf{y}}{\|\mathbf{x} - \mathbf{y}\|_2} , \frac{\mathbf{z} - \mathbf{y}}{\|\mathbf{z} - \mathbf{y}\|_2} \right\rangle.
\end{equation}
$f^e$ and $f^a$ denote Euclidean and angular distance functions, respectively. 
\( \mu \) represents the average distance among all tokens in the mini-batch and \( \langle \mathbf{x}, \mathbf{y} \rangle \) denotes the dot product of \( \mathbf{x} \) and \( \mathbf{y} \).

To measure the semantic relationships within the text embedding space of CLIP, we first gather [EOS] tokens of CLIP in the mini-batch. 
We define the set of [EOS] tokens in a mini-batch as follows:
\begin{equation}
    T^{\text{EOS}}=\{{T}_{1, L_q}, {T}_{2,L_q}, \dots, {T}_{B_q, L_q} \}\in\mathbb{R}^{B_q\times d_{\text{CLIP}}},
\end{equation}
where ${T}_{1, L_q}$ represents the [EOS] token of the first text query within the mini-batch.
Note that [EOS] is used for the distillation since [EOS] conveys more informative clues than other tokens in CLIP~\cite{yi2024towards} and using [EOS] reduces computational overhead compared to token-wise distillation.
Then, the knowledge of CLIP is distilled towards the encoded text tokens, $\bar{T}$.
Specifically, we distill the pairwise Euclidean distance relationships and triplet angular distance relationships from the CLIP text embeddings into the text-video joint embedding space.
The distillation process is expressed as:
\begin{equation}
    % \begin{split}
    \mathcal{L}^{\text{E}}\!=\!\frac{1}{B_q(B_q - 1)} \!\sum_{\substack{i, j \in \mathcal{B}_q \\ i \neq j}} \!\mathcal{L}^{\text{H}}\big(f^{\text{e}}(T^{\text{EOS}}_{i}, T^{\text{EOS}}_{j}), f^{\text{e}}(\bar{T}_{i}, \bar{T}_{j})\big),
    % \end{split}
\end{equation}
\begin{equation}
    % \begin{split}
    \!\!\mathcal{L}^{\text{A}}\!=\!\!\frac{1}{{B_q}^3}
    \!\!\sum_{\substack{i, j, k \in \mathcal{B}_q \\}} \!\!\!\mathcal{L}^{\text{H}}\big(f^{\text{a}}(T^{\text{EOS}}_{i}\!, T^{\text{EOS}}_{j}\!, T^{\text{EOS}}_{k}), f^{\text{a}}(\bar{T}_{i}, \bar{T}_{j}, \bar{T}_{k})\big),
    % &\sum_{\substack{i, j, k \in B_q \\ i \neq j, j \neq k, i \neq k}} \mathcal{L}_{\text{H}}\big(f_{\text{a}}(T_{i,L_q}, T_{j,L_q}, T_{k,L_q}), f_{\text{a}}(\bar{T}_{i}, \bar{T}_{j}, \bar{T}_{k})\big).
    % \end{split}
\end{equation}
where $\mathcal{B}_{q}=\{1,2,\ldots, B_q\}$ stands for a set of indices such that $|\mathcal{B}_{q}| = B_q$ and $\mathcal{L}^{\text{H}}$ denotes Huber loss~\cite{huber_loss}, which leads stable training by behaving as L2 loss for small errors and L1 loss for large errors.
Finally, the objective for TCPL is defined as follows: 
\begin{equation}
    \mathcal{L}^{\text{TCPL}} = \lambda^E\mathcal{L}^{\text{E}} + \lambda^A\mathcal{L}^{\text{A}},
\end{equation}
where $\lambda^E$ and $\lambda^A$ are weights for $\mathcal{L}^{\text{E}}$ and $\mathcal{L}^{\text{A}}$, respectively.
By preserving the well-structured semantic relationships within the foundation model, TCPL mitigates semantic collapse within text embeddings.

\subsection{Semantic Collapse in Video Embeddings: Cross-Branch Video Alignment}
\label{Sec.vr}
Semantic collapse also occurs within the video modality.
While the conventional text-video retrieval loss effectively pushes apart videos with different semantics, it does not explicitly preserve the multi-contextual nature of events within a single video.
As a result, contextually distinct segments within the same video may collapse into similar embeddings, limiting intra-video discriminability.

Therefore, we introduce Cross-Branch Video Alignment~(CBVA) that aims to disentangle the representations of distinct events within a video, thereby mitigating semantic collapse. 
Specifically, we leverage the representations from the typical dual-branch architecture used in PRVR frameworks, with separate encoders for clip- and frame-level branches~\cite{ms-sl_PRVR, gmm_PRVR}.
% Therefore, we leverage the representations from the typical dual-branch architecture used in PRVR frameworks, with separate encoders for clip- and frame-level branches~\cite{ms-sl_PRVR, gmm_PRVR}.
% Specifically, we introduce Cross-Branch Video Alignment~(CBVA) that aims to disentangle the representations of distinct events within a video. 
In CBVA, timestamp correspondence is leveraged to align each video frame with its matching clip segment while repelling it from segments at other timestamps.
% Specifically, we introduce Cross-Branch Video Alignment~(CBVA) that aligns the pairs of frame segments and their corresponding clip segments.
However, simply aligning different levels of video representation proves ineffective.
This issue stems from the common practice of generating clip segments by uniformly average‐pooling fixed-length segments~\cite{ms-sl_PRVR, gmmv2_PRVR}, which causes each clip to cover multiple contexts that can overlap across adjacent segments.

\textbf{Order-Preserving Token Merging.} To address the fragmentation of temporally adjacent content in untrimmed videos, we first introduce Order-Preserving Token Merging~(OP-ToMe) to construct consistent clip segments $V_c$, as shown in Fig.~\ref{fig:Overview}.
Unlike general token-merging schemes that may fuse tokens from arbitrary spatial or temporal locations~\cite{tome, tempme}, OP-ToMe restricts all merging operations to pairs of tokens drawn from successive frames, thereby preserving the original playback order~(for stable temporal modeling). 
% and preventing the collapse of semantically distinct events. 
Concretely, given a sequence of per-frame tokens, we first compute cosine similarities between disjoint adjacent-frame pairs.
% Concretely, given a sequence of per-frame tokens, we first compute cosine similarities between adjacent‐frame token pairs.
We then select the approximately top-$N\%$ of most similar adjacent‐frame pairs and merge each into a single clip token. 
This merging procedure is repeated for $M$ iterations until the frames are aggregated into the standard 32 clips used in prior work.
At each merge, the two tokens are fused via a size-weighted average of their feature vectors. 
Note that the proportional attention mechanism~\cite{tome} is integrated in our framework to account for each token's size~(the number of raw frames it represents).
% We then select the top-$N\%$ of most similar adjacent‐frame pairs and iteratively merge them over $M$ steps combined with proportional attention to account for each token’s size~(the number of raw frames it represents). 
By repeating this process, OP-ToMe produces a condensed sequence of clip segments that (1) maintain strict temporal order, (2) retain coherent contextual semantics, and (3) reduce redundant information across frames—properties that are crucial for robust performance in PRVR.
We provide the algorithm for OP-ToMe in the Appendix.
% temporal order는 frame과 매칭해야하니 중요.

% \paragraph{Clip-Frame Contrastive Alignment.}
\textbf{Cross-Branch Video Alignment.} Once the context-consistent clips are constructed via OP-ToMe, we perform cross-branch contrastive learning to encourage fine-grained temporal discriminability within each video.
Specifically, each clip token and its corresponding frame tokens are treated as positive pairs, while frame tokens from other temporal moments in the same video are regarded as negatives.
This facilitates the model in learning to distinguish between different contextual segments of a single video.
Formally, given that $\bar{V}_c = \{\bar{v}_{c}^{(i)}\}_{i=1}^{L_c}$ and $\bar{V}_f = \{\bar{v}_{f}^{(j)}\}_{j=1}^{L_f}$ denote the clip-level and frame-level video tokens respectively, we also define the set of associated frames of each clip $i$ as: \begin{equation}
    \mathbb{F}_i = \{\bar{v}_{f}^{j} |  \delta{(j)} = i \}, \quad X_i = \vert \mathbb{F}_i\vert,
\end{equation}
where $\delta(\cdot)$ returns the clip index of a frame among the $L_c$ clips. 
Then, the objective of CBVA is formulated with frame-to-clip and clip-to-frame NCE as:
\begin{equation}
\label{eq.vidcon}
 %        \!\!\!\mathcal{L}^{\text{CBVA}}\! =\! -\frac{1}{L_f}\sum_{i=1}^{L_f}\log\frac{\exp(\text{sim}(\bar{v}_{f}^{i}, v^{\delta{(i)}}_c))}{\sum_{j=1}^{L_c}\exp(\text{sim}(\bar{v}_{f}^{i}, \bar{v}_{c}^{j}))}
 % -\frac{1}{L_c}\sum_{i=1}^{L_c}\log\!\frac{\exp( \sum_{x=1}^{X_i}\text{sim}(\mathbb{F}_{i}[x], \bar{v}_{c}^{i}))}{\sum_{j=1}^{L_f}\exp(\sum_{x=1}^{X_i}\text{sim}(\mathbb{F}_{j}[x], \bar{v}_{c}^{i}))}), \\
 \!\!\!\mathcal{L}^{\text{CBVA}}\! =\! -\frac{1}{L_f}\sum_{i=1}^{L_f}\log\frac{\exp(\text{sim}(\bar{v}_{f}^{i}, v^{\delta{(i)}}_c))}{\sum_{j=1}^{L_c}\exp(\text{sim}(\bar{v}_{f}^{i}, \bar{v}_{c}^{j}))}
 -\frac{1}{L_c}\sum_{i=1}^{L_c}\log\!\frac{ \sum_{x=1}^{X_i}\exp(\text{sim}(\mathbb{F}_{i}[x], \bar{v}_{c}^{i}))}{\sum_{j=1}^{L_f}\exp(\text{sim}(\bar{v}_{f}^{j}, \bar{v}_{c}^{i}))}), \\
\end{equation}
where $\text{sim}(\cdot, \cdot)$ denotes cosine similarity and $\mathbb{F}_{i}[x]$ is the $x$-th frame token in the set $\mathbb{F}_{i}$.
% Formally, given that $\bar{V}_c = \{\bar{v}_{c}^{(i)}\}_{i=1}^{L_c}$ and $\bar{V}_f = \{\bar{v}_{f}^{(j)}\}_{j=1}^{L_f}$ denote the clip-level and frame-level visual tokens respectively, we also define the set of associated frames of each clip $i$ as: \begin{equation}
%     \mathbb{F}_i = \{\bar{v}_{f}^{j} |  \delta{(\bar{v}_{f}^{j})} = i \}, \quad X_i = \vert \mathbb{F}_i\vert,
% \end{equation}
% where $\delta(\cdot)$ returns the clip index of a frame among the $L_c$ clips. 
% Then, the objective of CBVA is formulated with frame-to-clip and clip-to-frame NCE as:
% \begin{equation}
% \label{eq.vidcon}
%         \!\!\!\mathcal{L}_{\text{CBVA}}\! =\! -\frac{1}{L_f}\sum_{i=1}^{L_f}\log\frac{\exp(\text{sim}(\bar{v}_{f}^{i}, \bar{v}_{c}^{p}))}{\sum_{j=1}^{L_c}\exp(\text{sim}(\bar{v}_{f}^{i}, \bar{v}_{c}^{j}))}
%  -\frac{1}{L_c}\sum_{i=1}^{L_c}\log\!\frac{\exp( \sum_{x=0}^{X}\text{sim}(\mathbb{F}_{i}[x], \bar{v}_{c}^{i}))}{\sum_{j=1}^{L_c}\exp(\sum_{x=0}^{X}\text{sim}(\mathbb{F}_{j}[x], \bar{v}_{c}^{i}))}), \\
% \end{equation}
% where $\text{sim}(\cdot, \cdot)$ denotes cosine similarity, $\bar{v}_{c}^{p}$ indicates the corresponding clip where each frame is included, and $\mathbb{F}_{i}[x]$ is the $x$-th frame token in the set $\mathbb{F}_{i}$.
% Also, for clip-to-frame NCE~(latter term), the positive frame set of $i$-th clip $\bar{v}_{c}^{i}$ is defined as $\mathbb{F}_i = \{\bar{v}_{f}^{i}\} \cup \{\bar{v}_{f}^{j} | j \in {L_f}, \delta{(\bar{v}_{f}^{j})} = \delta(\bar{v}_{f}^{i}) \}$.

\textbf{Adaptive CBVA.} 
Although CBVA disentangles different contexts within a single video, real-world footage often contains an unknown~(potentially variable) number of distinct contexts. 
Consequently, applying the contrastive objective in Eq.~\ref{eq.vidcon} with a fixed clip length $L_c$ may introduce noise: for example, an interview video composed of largely homogeneous frames will nonetheless be split into $L_c$ segments, unnecessarily fragmenting coherent content. 
To address this, we first estimate the number of contexts in each video and then adaptively aggregate $L_c^*$ representative clips to guide precise CBVA. 
We employ bipartite token merging~\cite{tome} to extract representative clip segments, since semantically similar content may occur intermittently or across non-contiguous intervals within a video.
However, optimizing the number of semantics per video is costly during the token merging process.
Therefore, we instead pre-define a discrete set of clip numbers based on a fixed merge rate, and then match each video to the level that best reflects its internal similarity structure~(number of different semantics).
To initially establish a discrete set of clip levels, we define $N\%$ to denote the merge rate and $C_\text{min}$ to represent the minimum number of semantically different clips in each video.
Then, we generate $K$ levels of clip number candidates $\{L_c^i\}_{i=1}^{K}$ by recording clip number after each merge step as:
%However, directly optimizing the context count per video is costly, so we instead define a discrete set of clip levels based on a fixed merge rate and then match each video to the level that best reflects its internal similarity structure.
%Let $N\%$ denote the merge rate and $C_\text{min}$ the minimum clip count. 
%We generate $K$ levels of clip count candidates $\{L_c^i\}^{K}_{i=1}$ by recording clip counts after each merge step as:
\begin{equation}
\label{eq.cliplevel}
L_c^1 = L_c, \quad L_c^{i+1} = \max\bigl(2 \times \lfloor  \frac{L_c^i - (L_c^i / 2) \times (N / 100) + 1}{2}   \rfloor,\,C_{\min}\bigr),
\end{equation}
and let $K$ be the largest index for which $L_c^K \ge C_{\text{min}}$.
Next, we compute a high-similarity ratio $\omega$ for each video by measuring the fraction of clip-pair cosine similarities~(using frozen features from the backbone $V_c$) that exceed a threshold $\tau$.
A low $\omega$ indicates many distinct contexts, so we retain the full original clip set (${L}_c^*=L_c$).
Otherwise, we select the smallest $k \in \{1,\ldots, K\}$ satisfying $\omega > \frac{K-k}{K}$, and perform $k\!-\!1$ iterations of bipartite merging at rate $N\%$, yielding $L_c^* = L_c^k$ final clips.
We remark that, for simplicity, we use the same merge rate $N\%$ as OP-ToMe. 
Consequently, in Eq.~\ref{eq.vidcon}, the original clip segments are replaced with these merged clips to further enhance video adaptivity.
Detailed algorithm for both merging processes are provided in the Appendix.

\subsection{Total Training Objective}
\label{Sec.objective}
Finally, our total objective with retrieval, TCPL, and CBVA losses is expressed as:
\begin{equation}
    \mathcal{L}^\text{overall} = \mathcal{L}^\text{base} + \mathcal{L}^\text{TCPL} + \lambda^{\text{CBVA}}\mathcal{L}^{\text{CBVA}}.
\end{equation}

%% file: sec/4_experiments.tex
\input{table/tab0}

\section{Experiments}
\label{sec.experiments}

% \subsection{Experimental Setting}
\textbf{Datasets \& Metrics.}
We evaluated our method on four PRVR datasets: QVHighlights~\cite{qvhighlight}, TVR~\cite{tvr_DATA}, ActivityNet Captions~\cite{anetcaptions_DATA}, and Charades-STA~\cite{charades_DATA}. 
QVHighlights\cite{qvhighlight} is a collection of news and vlog-style videos, recently reorganized for PRVR\cite{protoprvr}. 
Each video is paired with an average of 3.3 text queries describing semantically diverse segments.
TVR~\cite{tvr_DATA} is built from scenes across six popular TV shows, with each video annotated by five text queries targeting different segments. 
The training set contains 17,435 videos and 87,175 queries, while the evaluation set includes 2,179 videos and 10,895 queries.
ActivityNet Captions~\cite{anetcaptions_DATA} is sourced from YouTube videos, with an average of 3.7 text queries per video. 
The dataset includes 10,009 videos for training and 4,917 for evaluation.
Charades-STA~\cite{charades_DATA} extends the original Charades dataset by adding sentence-level annotations for specific temporal segments. 
It consists of 13,898 video-sentence pairs for training and 4,233 for evaluation.
% Details for datasets are in Appendix.
% We use re-organized QVHighlights~\cite{protoprvr}, which consists of 7,218 partially relevant video-text pairs in the training set and 1,550 pairs in the test set.
% TVR~\cite{tvr_DATA} is a dataset constructed based on scenes from six TV shows, where each video is assigned five textual queries describing different segments of the video. The training set consists of 17,435 videos and 87,175 text queries, while the evaluation set comprises 2,179 videos and 10,895 text queries. 
% ActivityNet Captions~\cite{anetcaptions_DATA} is a dataset constructed from YouTube videos. 
% On average, each video is assigned 3.7 text queries. The training set consists of 10,009 videos, while the evaluation set comprises 4,917 videos.
% The Charades-STA~\cite{charades_DATA} dataset is an extension of the original Charades dataset, augmented with textual descriptions of specific segments within the videos. The training set consists of a total of 13,898 video-sentence pairs, while the evaluation set contains 4,233 pairs.
% \textbf{Evaluation Metrics}
For evaluation, we use recall-based metrics, which are commonly used in retrieval tasks~\cite{img_retr, t2vr, asym_img_retr, semi_img_retr, uatvr, tmass}. 
We denote this metric as R@$Q$, where $Q$ represents the proportion of queries for which the correct video appears within the top-$Q$ ranked results. 
% We evaluate this metric at various levels of K, specifically 1, 5, 10, and 100. 
Additionally, SumR is the sum of all R@$Q$ used for evaluation, assessing the overall retrieval performance.
% Additionally, to comprehensively assess retrieval performance across different levels, we also report SumR, which is the sum of these recall values, following previous approaches~\cite{gmm_PRVR,gmmv2_PRVR,ms-sl_PRVR}.

\textbf{Implementation Details.}
% We follow prior works in choosing backbones.
% Specifically, for QVHighlights, we use CLIP-B/16~\cite{clip_ID} and Slowfast~\cite{feichtenhofer2019slowfast} for encoding visual and text features.
% For TVR, ActivityNet-Captions, and Charades, we use CLIP-L/14 following~\cite{qasir,, protoprvr}.
% For feature extraction, we follow recent works; we use CLIP-B and Slowfast for QVHighlights, and CLIP-L for other datasets.
For feature extraction, we follow recent works~\cite{arl, qasir, protoprvr}; we extract video features with CLIP-B/16~\cite{clip_ID} and Slowfast~\cite{feichtenhofer2019slowfast}, and use CLIP-B for text embeddings for QVHighlights, and use CLIP-L~\cite{clip_ID} for encoding both modalities in other datasets.
Hyperparameter configurations are adopted from GMMFormer-v2~\cite{gmmv2_PRVR}~(e.g., learning rate, batch size, epochs, and optimizer settings) except for the fusing ratio between the frame and clip branches.
We assign a frame score weight of 0.6 and a clip score weight of 0.4.
All loss coefficients are fixed across datasets: $\lambda^E = 15$, $\lambda^A = 30$, and $\lambda^{\text{CBVA}} = 0.1$.
To construct consistent clips with OP-ToMe, we set $N$ to 75\%~(Note that $M$ is then computed automatically from $N$ to match the number of clips used in prior works~\cite{gmmv2_PRVR, ms-sl_PRVR}.)
Finally, we set the minimum clip count per video to $C_\text{min} = 5$, and set a similarity threshold $\tau$ to 0.7 for QVHighlights, 0.8 for TVR and ActivityNet-Captions, and 0.85 for Charades.
The reason behind using varying $\tau$ is that the internal segment-to-segment similarity distributions differ; QVHighlights exhibits the lowest similarities, TVR and ActivityNet-Captions are intermediate, and Charades shows the highest.
All experiments are conducted on a single RTX A6000 GPU and an Intel Xeon Gold 6338 CPU (2.00GHz) for all datasets.

% adaptive CBVA.

\subsection{Ablation Study}
Studies are conducted on QVHighlights, which includes numerous events in each untrimmed video.
The default configuration used to generate the reported results is highlighted in grey.

\noindent\textbf{Component ablation.}
To quantify the contribution of each module, we report a component‐wise ablation in Tab.~\ref{Tab.Ablation1}.
Our baseline is built upon GMMFormer-v2 architecture~\cite{gmmv2_PRVR}, only trained with the standard retrieval loss $\mathcal{L}^{\text{base}}$.
Then, we sequentially add Text Correlation Preservation learning~(TCPL) and Cross-Branch Video Alignment~(CBVA), which are introduced in Sec.~\ref{Sec.tcpl} and Sec.~\ref{Sec.vr}.
Initially, in row (b), incorporating TCPL mitigates semantic collapse in the text embedding space, yielding a notable gain over the baseline.
From row (c) to (e), we subdivide the CBVA into (c) Na\"ive CBVA, (d) adding OP-ToMe, and (e) applying adaptive CBVA.
Specifically, the basic CBVA objective produces only a marginal increase in performance since fixed-length clip segments may encompass multiple overlapping contexts.
However, we find that augmenting CBVA with OP-ToMe to construct semantically consistent clip segments drives a performance boost by reducing spurious alignments across events.
Finally, dynamically adjusting each video's clip count according to the estimated number of video contexts further refines the alignment, producing a substantial gain.
These results confirm that addressing both the text- and video-side semantic collapse is significant for PRVR.

\textbf{Video Correlation Preservation Learning~(VCPL).}
Similar to TCPL, one can assume that we can apply the identical approach to video embeddings to mitigate semantic collapse.
However, this direct adaptation is suboptimal since CLIP's video embeddings cannot model temporal dynamics.
To substantiate this, Tab.~\ref{tab:VideoCPL} compares VCPL against our CBVA. 
`Retrieved segment' is conducted similarly to TCPL; we first select the representative video token for every text query by identifying the token with the highest similarity within the paired videos~(using ground-truth pair) and distill the relationships between representative video segments.
% `Retrieved segment' in the first row indicates that VCPL is conducted similarly to TCPL; we used the retrieved video segments from the paired videos for all text queries~(using ground-truth pair information) and distilled the relationships between these video segments.
Also, we study the variant of VCPL where we uniformly sample 4 segments per video and conduct relation learning between all sampled segments from the mini-batch.
Although these approaches yield a modest improvement, we find that these variants lag behind CBVA by 2.3 points in SumR.
VCPL is applied to both clip and frame branches.

\input{table/tab_ablation_various}
\noindent\textbf{Loss coefficients.}
For our training objective, we control the TCPL loss with $\lambda^E$ and $\lambda^A$, and the CBVA loss with $\lambda^{\text{CBVA}}$.
In Tab.~\ref{Tab.rkd1}, we first studied the $\lambda^E$ : $\lambda^A$ over $\{1:1, 2:1, 1:2\}$.
Then, in Tab.~\ref{Tab.rkd2} with a 1:2 ratio, which yields the best performance, increasing both weights to (15, 30) improved performance; beyond that, gains plateaued.
For CBVA, in Tab.~\ref{Tab.rkd6}, performance rose as $\lambda^{\text{CBVA}}$ increased up to 0.15, but for simplicity across datasets, we fixed it at 0.1.

\noindent\textbf{TCPL source model.}
By default, we use the pretrained text encoder as the source model for TCPL to provide semantic relationships~(CLIP-B for QVHighlights and CLIP-L for other datasets).
% By default, we use the CLIP-B model as the source model for TCPL to provide semantic relationships.
To assess sensitivity to the source model, we replaced CLIP-B with alternative vision–language encoders and measured SumR on the QVHighlights dataset in Tab.~\ref{Tab.rkd3}. 
As observed, swapping in the larger models~(e.g., CLIP-L and OpenCLIP-L) increased SumR by up to 1.8 points.
These results indicate that TCPL’s effectiveness scales with the quality of the source model’s semantic structure.

\noindent\textbf{Token-Merging Ratio.}
We use a single merge rate $N$\% for both OP-ToMe and adaptive CBVA.
Empirically, setting $N$ to approximately 75\% reduces 128 frames to 32 clips in only a few steps~(matching the standard PRVR frame/clip counts), while keeping computational overhead minimal.
As Tab.~\ref{Tab.rkd4} shows, increasing the number of merge iterations while lowering the per-step ratio to 50\% actually degraded accuracy.
Thus, we fix $N = 75\%$ across all datasets.

\noindent\textbf{Adaptively measuring video context number.} 
We determine the optimal number of contexts for each video by thresholding the pairwise similarity among its clips at a value $\tau$. 
In this work, we vary $\tau$ to evaluate how sensitive our context-count estimation is to this threshold. 
As shown in Tab.~\ref{Tab.rkd5}, the adaptive CBVA method exhibits only minor fluctuations across different $\tau$ values, indicating that it is robust to the choice of similarity threshold between 0.5 and 0.8.

\subsection{Comparison with the State-of-the-Art}\input{table/tab6_videorkd}
\textbf{QVHighlights.}
In Tab.~\ref{Tab.qv}, we report results on QVHighlights~\cite{qvhighlight}, a recently introduced benchmark for PRVR.
To illustrate, our method outperforms the previous state of the art by up to 8 points in SumR.
We attribute these gains to our method's capability to mitigate semantic collapse, especially when videos exhibit frequent and rapid event transitions.
\input{table/tab1_performance}

\textbf{TVR} \& \textbf{ActivityNet-Captions} \& \textbf{Charades.}
Tab.~\ref{table_tvr} reports results on these three datasets. 
Specifically, our method achieves state-of-the-art results on all datasets.
The performance gains on these datasets are relatively modest compared to QVHighlights, primarily because QVHighlights exhibits very little overlap between different queries and video segments for the same video, making it especially susceptible to semantic collapse.
Despite this, our method maintains state-of-the-art performance across all benchmarks, underscoring its generalizability and effectiveness.

\begin{table}
\centering
\vspace{-0.2cm}
\footnotesize
\caption{Inference time~(ms) and memory~(MB) across varying size of video database.}
\label{Tab.efficiency}
\setlength{\tabcolsep}{12.4pt}
\renewcommand{\arraystretch}{0.85}
\begin{tabular}{llccccc}
\toprule
\multirow{2}{*}{Method} & \multirow{2}{*}{Metric} & \multicolumn{5}{c}{Number of Videos} \\
\cmidrule(lr){3-7}
 & & 100 & 200 & 300 & 400 & 474 \\
\midrule
\multirow{2}{*}{MSSL}
 & Time (ms)    & 3.09 & 3.85 & 4.66 & 5.14 & 5.58 \\
 & Memory (MB)  & 717.47 & 796.15 & 874.83 & 954.14 & 1010.89 \\
\midrule
\multirow{2}{*}{GMMF}
 & Time (ms)    & 1.97 & 1.98 & 1.99 & 2.02 & 2.05 \\
 & Memory (MB)  & 243.11 & 248.95 & 254.78 & 260.62 & 264.10 \\
\midrule
\multirow{2}{*}{GMMF-v2}
 & Time (ms)    & 2.31 & 2.38 & 2.40 & 2.61 & 2.78 \\
 & Memory (MB)  & 419.75 & 440.18 & 459.62 & 480.55 & 493.46 \\
\midrule
\multirow{2}{*}{Ours}
 & Time (ms)    & 2.32 & 2.37 & 2.40 & 2.60 & 2.70 \\
 & Memory (MB)  & 419.75 & 440.18 & 459.62 & 480.55 & 493.46 \\
\bottomrule
\vspace{-0.5cm}
\end{tabular}
\end{table}
\begin{wraptable}{h!}{0.52\textwidth}
\centering
\footnotesize
\vspace{-0.45cm}
\setlength{\tabcolsep}{2.pt}
\renewcommand{\arraystretch}{0.9}
\caption{Training efficiency and model complexity.}
\label{Tab.trainingdetail}
\vspace{-0.cm}
    \begin{tabular}{lrrrr}
    \toprule
    Training details & MSSL & GMMF & GMMF-v2 & Ours \\
    \midrule
    Time / epoch (ms)      & 10,934 & 12,828 & 17,223 & 62,641 \\
    Memory (MB)            &  2,375 &  3,333 &  7,826 &  9,755 \\
    Model params (M)   &   4.57 &  12.72 &  32.14 &  32.14 \\
    FLOPs (G)              &   0.37 &   0.99 &   2.78 &   2.78 \\
    \bottomrule
    \end{tabular}
\end{wraptable}
\textbf{Efficiency.}
In Tab.~\ref{Tab.efficiency},~\ref{Tab.trainingdetail}, we report inference/training time and memory, along with model parameters and FLOPs on QVHighlights.
Reported times are averaged over 5 runs.
For the inference, we measure the inference time and memory across database sizes from 100 to 474 videos.
As shown, our method attains the second-lowest inference latency and memory footprint while achieving substantially higher retrieval accuracy.
Note that inference time refers to query time since video features are precomputed and cached in practical deployments.
Training statistics in Tab.~\ref{Tab.trainingdetail} show higher time and memory due to learning fine-grained video context, but this cost is paid offline, whereas inference efficiency governs real-world deployment where latency and memory are critical.

% \subsection{Further Study}
% \paragraph{rkd a혹은 d 둘중 하나만 쓰면?}
% \paragraph{Cross-Domain}

\subsection{Analysis}
\input{table/tab3_sentence_sim}
\textbf{Similarity Structure.}
We compare the pairwise similarity between queries~(video segments) associated with the same video~(\textit{Intra Sim}) and between all instances across videos~(\textit{Total Sim}). 
If the relationship between contexts and their descriptive queries within each video were indistinguishable from that observed across different videos, \textit{Diff. Norm} would equal 0; if every context within a video were identical, \textit{Diff. Norm} would equal 1. 
For the analysis, we leverage QVHighlight to assess semantic collapse via similarity structure, as it exhibits relatively minimal semantic overlap among queries within the same video. 
As shown in Tab.~\ref{tab:semantic_collapse_sim}, our method substantially reduces \textit{Diff. Norm} to a point where we claim that our method preserves an appropriate level of relative coherence within each video~(not too low) while also mitigating semantic collapse~(not too high).

% We remark that Diff.Norm of CLIP i

% \textbf{Impact of Order-Preserving Token Merging.}
% Furthermore, in Tab.~\ref{tab:semantic_collapse_sim}, we check whether OP-ToMe 
% - token merge 했을 때 clip pretrained feature similarity가 바뀐다는 분석 추가필요
% - 토큰 간에 similarity가 줄어들어서 비슷한 context끼리 뭉쳐서 토큰들은 다른 context가 됨.
% - 고로 OP-ToMe의 효과가 확인됨.

\input{table/tab4_inter_intra_text_sim}
\textbf{Spearman rank correlation with CLIP.}
We assess whether our method effectively preserves the semantic structure compared to baseline approaches. 
Specifically, we measure how each method preserves the semantic structure of CLIP using Spearman's rank correlation~\cite{spearman}.
For the evaluation, we use the pooled text tokens $\bar{T}$ from each PRVR model to compare with the [EOS] tokens within CLIP query embeddings.
Tab.~\ref{tab:spearman} demonstrates how our proposed method well preserves the semantic relationships between text queries, thereby mitigating semantic collapse.

\begin{wrapfigure}{r}{0.49\textwidth}
    \centering
    \vspace{-0.5cm}
    \includegraphics[width=0.48\textwidth]{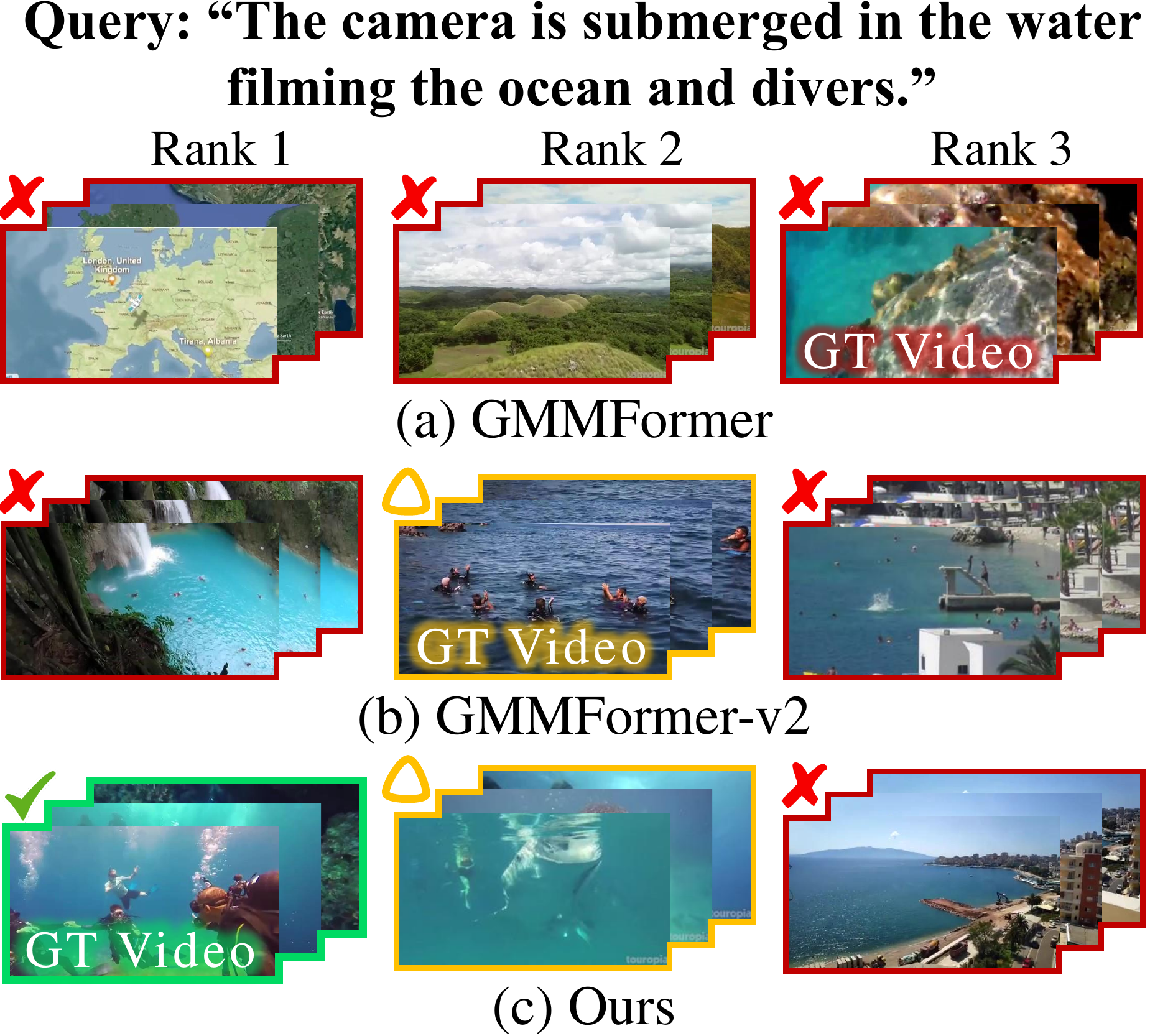}
    \vspace{-0.1cm}
    \caption{
        Retrieval example. 
        % The video at Rank n corresponds to the n-th retrieved result for a given text query. 
        `GT Video' denotes the ground-truth paired video to the query. 
        $\checkmark$, $\triangle$, and \ding{55} indicate whether the retrieved video token is semantically aligned or not, regardless of its origin from the ground-truth video.
        % $\checkmark$ indicates whether the retrieved video token is semantically aligned with the query, regardless of 
        % Similarly, $\triangle$ indicates partial semantic alignment, while \ding{55} denotes a complete semantic mismatch. 
        % We evaluate our method in comparison with GMMF and GMMFormer-v2.
    }
    \label{fig:Qualitative}
    \vspace{-1.5cm}
\end{wrapfigure}
\textbf{Qualitative results.}
Fig.~\ref{fig:Qualitative} shows qualitative retrieval results for a text query. 
Our method correctly retrieves and localizes the video token that overlaps the query’s target moment~(within additional temporal margin~\cite{zhao2017temporal, lin2018bsn, protoprvr}), whereas the baseline models are distracted by superficially similar content~(depicting generic ocean scenes).
This failure stems from their embedding collapse, which blurs distinct events with similar global semantics. 
In contrast, by preserving fine‐grained semantic structure, our approach disambiguates these contextually similar contexts and retrieves the exact segment corresponding to the query.
% fig3은 retrieval이 개선된 case에 대한 시각화이다.
% baseline은 query에 대한 visual semantic align에 어려움을 겪고 있음 (조금더 자세히 쓸 필요 있나?)
% 그 대신에 우리꺼는 대체적으로 쿼리의 내용을 담고 있는 비디오 클립을 잘 찾고 있으며 이에 따라 정답 비디오도 잘 맞춤
% 결론적으로, 우리가 제안하는 방법들은 semantic collapse를 완화하여 model이 semantic을 더 잘 이해하여 관련된 비디오들을 더 잘 찾을 수 있음 (prvr에 도움이 된다.)
% fig3은 baseline 모델에 비해 retrieval이 개선된 case에 대한 시각화이다.
% FIG3 A AND B에서 보이듯이, 기존 방법들은 수중 장면이 아니거나 다이버가 없는 등 쿼리와 관련없는 프레임을 통해 비디오를 찾고 있다.
% 이 때문에 GT 비디오를 찾는 데 어려움을 겪고 있어, GT 비디오의 ranking이 내려간다. 
% 하지만 Fig3 (c)에서 보이듯이 ours는 쿼리의 문장 전체의 의미와 일치하는 프레임을 정확히 찾아낸다. 
% 이로 인해, 우리의 방법은 GT 비디오를 정확히 찾을 수 있으며 그 하위 랭킹의 비디오도 Semantic하게 유사한 것을 확인할 수 있다.
% 결과적으로 우리는 우리의 방법이 비디오나 text의 context 정보를 잘 보존하여 PRVR task 수행하는 데 도움을 줄 수 있다고 주장한다.

%% file: table/tab0.tex
\begin{table}[!t]
\vspace{-0.cm}
    \centering
    \footnotesize
    \begin{minipage}[t]{0.51\linewidth}%\centering
    \setlength{\tabcolsep}{2.2pt} % Default value: 6pt
    \renewcommand{\arraystretch}{0.95}
    \caption{Ablation study on QVHighlights dataset.}
    \label{Tab.Ablation1}
    \vspace{0.35cm}
    \centering
    \footnotesize
    \begin{tabular}{l|l|ccccc}
        \hlineB{2.5}
         & Model & R1 & R5 & R10 & R100 & SumR \\
        \hline \hline
        (a) & Baseline & 21.8 & 48.1 & 60.6 & 95.0 & 225.5 \\
        (b) & + TCPL & 22.8 & 49.5 & 63.3 & 95.0 & 230.6\\
        (c) & + Na\"ive CBVA & 22.8 & 49.4 & 63.7 & 95.0 & 231.0 \\
        (d) & + OP-ToMe & 24.2 & 50.4 & 63.0 & 94.9 & 232.5 \\ \rowcolor{gray!20}
        (e) & + Adaptive CBVA & 23.9 & 51.5 & 63.7 & 95.5 & 234.6 \\
        \hlineB{2.5}
    \end{tabular}
    % \caption{Correlation preservation learning on video embeddings.}
    % \label{Tab.cpl}
    % \centering
    % \footnotesize
    % \begin{tabular}{l|ccccc}
    %     \hlineB{2.5}
    %     Model & R@1 & R@5 & R@10 & R@100 & SumR \\
    %     \hline \hline
    %     TCPL & 22.8 & 49.5 & 63.3 & 95.0 & 230.6\\ \hline
    %     TCPL + VCPL.max & 23.4 & 50.4 & 63.4 & 94.6 & 231.7 \\
    %     TCPL + VCPL.samp & 22.5 & 50.8 & 64.1 & 94.9 & 232.3 \\
    %     TCPL + Adaptive CBVA  & 23.9 & 51.5 & 63.7 & 95.5 & 234.6 \\
    %     \hlineB{2.5}
    % \end{tabular}
    \end{minipage}\hfill
    \begin{minipage}[t]{0.48\linewidth}\centering
    \renewcommand{\arraystretch}{0.95}
    \setlength{\tabcolsep}{1pt} 
      \caption{Performance when using variants of video correlation preservation learning instead of Cross‐Branch Video Alignment.
  }
  \vspace{-0.05cm}
  \label{tab:VideoCPL}
  \setlength{\tabcolsep}{1.4pt}
  \renewcommand{\arraystretch}{.95}
  \begin{tabular}{l|ccccc}
    \hlineB{2.5}
    Method & R1 & R5 & R10 & R100 & SumR  \\ \hline\hline
    (a) TCPL baseline & 22.8 & 49.5 & 63.3 & 95.0 & 230.6 \\
    (a)+ Retrieved segment & 23.4 & 50.4 & 63.4 & 94.6  & 231.7 \\
    (a)+ Uniform Sampling & 22.5 & 50.8 & 64.1 & 94.9  & 232.3 \\ \rowcolor{gray!20}
    Ours & 23.9 & 51.5 & 63.7 & 95.5  & 234.6 \\ \hlineB{2.5}
  \end{tabular}
        % \caption{
        %     Results on QVHighlights. $\dagger$ denotes reproduced results.
        % }
        % \label{Tab.qv}
        % \begin{tabular}{c|ccccc}
        % \hlineB{2.5}
        % Methods & R@1 & R@5 & R@10 & R@100 & SumR \\  \hline \hline %
        %  % \cline{1-1} \cline{3-6} \cline{8-11} 
        % MS-SL~\cite{ms-sl_PRVR} & 20.4 & 46.7 & 60.7 & 94.6 & 222.5 \\
        % % DL-DKD~\cite{dldkd_PRVR}$\dagger$ & 20.5 & 44.5 & 55.8 & 92.8 & 213.7 \\
        % GMMF~\cite{gmm_PRVR} & 18.2 & 43.7 & 56.7 & 92.5 & 211.1 \\ 
        % GMMF-v2~\cite{gmmv2_PRVR}$\dagger$ & 21.7 & 48.0 & 60.5 & 95.0 & 225.2 \\ 
        % ProtoPRVR~\cite{protoprvr} & 22.6 & 48.8 & 61.3 & 93.9 & 226.6 \\ \rowcolor{gray!20} 
        % \textbf{Ours} & \textbf{23.9} & \textbf{51.5} & \textbf{63.7} & \textbf{95.5} & \textbf{234.6} \\  \hlineB{2.5}
        % \end{tabular}
    \end{minipage}
    \vspace{-0.4cm}
\end{table}

%% file: table/tab_ablation_various.tex
\begin{table}[!t]
\vspace{-0.cm}
  \centering
  \scriptsize
  \caption{Ablation studies of various components on QVHighlights. `Coef' denotes coefficient.}
  \vspace{-0.1cm}
  \label{Tab.ablation_various}
  \begin{subtable}[t]{0.16\linewidth}
    \centering
    \scriptsize
    \caption{TCPL ratio.}
    \label{Tab.rkd1}
    \vspace{-0.1cm}
    \setlength{\tabcolsep}{3pt}
    \renewcommand{\arraystretch}{0.9}
    \begin{tabular}{c|c}
      \hlineB{2.5}
      $\lambda^E:\lambda^A$ & SumR \\ \hline\hline
      1:1~(15,15) & 229.7 \\
      2:1~(30,15) & 231.8 \\ \rowcolor{gray!20}
      1:2~(15,30) & 234.6 \\ \hlineB{2.5}
    \end{tabular}
  \end{subtable}\hfill
  \begin{subtable}[t]{0.16\linewidth}
    \centering
    \scriptsize
    \caption{TCPL coef.}
    \label{Tab.rkd2}
    \vspace{-0.1cm}
    \setlength{\tabcolsep}{3.3pt}
    \renewcommand{\arraystretch}{0.9}
    \begin{tabular}{cc|c}
      \hlineB{2.5}
      $\lambda^E$ & $\lambda^A$ & SumR \\ \hline\hline
      5           & 10          & 231.5 \\
      10          & 20          & 233.5 \\ \rowcolor{gray!20}
      15          & 30          & 234.6 \\
      20          & 40          & 232.5 \\ \hlineB{2.5}
    \end{tabular}
  \end{subtable}\hfill
  \begin{subtable}[t]{0.20\linewidth}
    \centering
    \scriptsize
    \caption{TCPL Source.}
    \label{Tab.rkd3}
    \vspace{-0.1cm}
    \setlength{\tabcolsep}{4pt}
    \renewcommand{\arraystretch}{0.9}
    \begin{tabular}{c|c}
      \hlineB{2.5}
      Model          & SumR \\ \hline\hline \rowcolor{gray!20}
      CLIP-B      & 234.6 \\ 
      CLIP-L      & 235.6 \\
      OpenCLIP-B  & 235.4 \\
      OpenCLIP-L  & 236.4 \\ \hlineB{2.5} 
      % 결과 전부 41번서버에 있음
    \end{tabular}
  \end{subtable}%\hfill
    \begin{subtable}[t]{0.15\linewidth}
    \centering
    \scriptsize
    \caption{CBVA coef.}
    \label{Tab.rkd6}
    \vspace{-0.1cm}
    \setlength{\tabcolsep}{5pt}
    \renewcommand{\arraystretch}{0.9}
    \begin{tabular}{c|c}
      \hlineB{2.5}
      $\lambda^{\text{CBVA}}$ & SumR \\ \hline\hline \rowcolor{gray!20}
      0.1    & 234.6 \\ 
      0.15    & 234.9 \\ 
      0.2    & 232.9 \\ 
      \hlineB{2.5}
    \end{tabular}
  \end{subtable}%\hfill
  \begin{subtable}[t]{0.15\linewidth}
    \centering
    \scriptsize
    \caption{Merge rate.}
    \label{Tab.rkd4}
    \vspace{-0.1cm}
    \setlength{\tabcolsep}{5pt}
    \renewcommand{\arraystretch}{0.9}
    \begin{tabular}{c|c}
      \hlineB{2.5}
      $N\%$ & SumR \\ \hline\hline
      50    & 232.6 \\ \rowcolor{gray!20}
      75    & 234.6 \\ \hlineB{2.5}
    \end{tabular}
  \end{subtable}%\hfill
  \begin{subtable}[t]{0.16\linewidth}
    \centering
    \scriptsize
    \caption{Threshold $\tau$.}
    \label{Tab.rkd5}
    \vspace{-0.1cm}
    \setlength{\tabcolsep}{5pt}
    \renewcommand{\arraystretch}{0.9}
    \begin{tabular}{c|c}
      \hlineB{2.5}
      $\tau$ & SumR \\ \hline\hline
      0.5    & 234.3 \\
      0.6    & 233.5 \\ \rowcolor{gray!20}
      0.7    & 234.6 \\
      0.8    & 232.6 \\ \hlineB{2.5}
    \end{tabular}
  \end{subtable}
  \vspace{-0.55cm}
\end{table}

%% file: table/tab6_videorkd.tex
\begin{wraptable}{tr}{0.45\textwidth}
  \vspace{-0.75cm}
  \centering
  \footnotesize
  % \caption{Performance when using variants of video correlation preservation learning instead of Cross‐Branch Video Alignment.
  % }
  % \vspace{-0.15cm}
  % \label{tab:VideoCPL}
  \setlength{\tabcolsep}{2.2pt}
  \renewcommand{\arraystretch}{1.}
  % \begin{tabular}{l|ccccc}
  %   \hlineB{2.5}
  %   Method & R1 & R5 & R10 & R100 & SumR  \\ \hline\hline
  %   TCPL baseline & 22.8 & 49.5 & 63.3 & 95.0 & 230.6 \\
  %   Retrieved segment & 23.4 & 50.4 & 63.4 & 94.6  & 231.7 \\
  %   Uniform Sampling & 22.5 & 50.8 & 64.1 & 94.9  & 232.3 \\ \rowcolor{gray!20}
  %   \textbf{Ours} & 23.9 & 51.5 & 63.7 & 95.5  & 234.6 \\ \hlineB{2.5}
  % \end{tabular}
        \caption{
            Results on QVHighlights. $\dagger$ denotes reproduced results.
        }
        \vspace{-0.2cm}
        \label{Tab.qv}
        \begin{tabular}{c|ccccc}
        \hlineB{2.5}
        Methods & R1 & R5 & R10 & R100 & SumR \\  \hline \hline %
         % \cline{1-1} \cline{3-6} \cline{8-11} 
        MS-SL~\cite{ms-sl_PRVR} & 20.4 & 46.7 & 60.7 & 94.6 & 222.5 \\
        % DL-DKD~\cite{dldkd_PRVR}$\dagger$ & 20.5 & 44.5 & 55.8 & 92.8 & 213.7 \\
        GMMF~\cite{gmm_PRVR} & 18.2 & 43.7 & 56.7 & 92.5 & 211.1 \\ 
        AMDNet~\cite{amdnet} & 17.4 & 40.8 & 55.0 & 93.4 & 206.6 \\ 
        BGMNet~\cite{BGMnet} & 20.6 & 46.3 & 58.8 & 94.0 & 219.7 \\ 
        GMMF-v2~\cite{gmmv2_PRVR}$\dagger$ & 21.7 & 48.0 & 60.5 & 95.0 & 225.2 \\ 
        ProtoPRVR~\cite{protoprvr} & 22.6 & 48.8 & 61.3 & 93.9 & 226.6 \\ 
        \rowcolor{gray!20} 
        \hline
        \textbf{Ours} & \textbf{23.9} & \textbf{51.5} & \textbf{63.7} & \textbf{95.5} & \textbf{234.6} \\  \hlineB{2.5}
        \end{tabular}
  \vspace{-0.5cm}
\end{wraptable}

%% file: table/tab1_performance.tex
\begingroup
\setlength{\tabcolsep}{2.2pt} % Adjust column spacing
\renewcommand{\arraystretch}{1.00} % Adjust row height (default is 1.0)
\begin{table*}[t!]
\vspace{-0.2cm}
\caption{
    Performances on TVR, ActivityNet Captions, and Charades-STA using CLIP-L/14 backbone.
    $\dagger$ are reproduced results, and all results on Charades are reproduced with official codes.
}
\label{table_tvr}
\vspace{-0.1cm}
\centering
\footnotesize % Reduced font size by 10%
\begin{tabular}{l|ccccc|ccccc|ccccc}
\hlineB{2.5}
\multirow{2}{*}{Method}
  & \multicolumn{5}{c|}{TVR}
  & \multicolumn{5}{c|}{ActivityNet Captions}
  & \multicolumn{5}{c}{Charades-STA} \\
\cline{2-16}
  & R1  & R5  & R10 & R100 & SumR
  & R1  & R5  & R10 & R100 & SumR
  & R1  & R5  & R10 & R100 & SumR \\
\hline
\hline
CLIP zero-shot
  & 16.2 & 33.5 & 41.8 & 75.7  & 167.2
  & 15.1 & 33.9 & 45.1 & 78.9  & 172.9
  & 2.0  & 8.1  & 13.6 & 49.4  & 73.1 \\
\hline
MS-SL~\cite{ms-sl_PRVR}
  & 31.9 & 57.6 & 67.7 & 93.8  & 251.0
  & 14.7 & 37.1 & 50.4 & 84.6  & 186.7
  & \textbf{3.4} & 11.5 & 18.7 & 62.5  & 96.0 \\
GMMF~\cite{gmm_PRVR}
  & 29.8 & 54.2 & 64.6 & 92.5  & 241.1
  & 15.2 & 37.7 & 50.5 & 83.7  & 187.1
  & 2.7  & 10.5 & 16.7 & 59.4  & 89.3 \\
AMDNet~\cite{amdnet}
  & 27.7 & 52.3 & 63.3 & 92.3 & 235.6
  & 14.0 & 36.3 & 49.9 & 84.2 & 184.5
  & 2.1 & 7.8 & 13.9 & 57.2 & 81.1 \\
BGM-Net~\cite{BGMnet}
  & 31.1 & 56.3 & 66.5 & 93.8 & 247.7
  & 15.6 & 37.9 & 51.3 & 85.4 & 190.3
  & 3.0 & 11.8 & 18.2 & 63.7 & 96.7 \\
GMMF-v2~\cite{gmmv2_PRVR}$\dagger$
  & 34.0 & 59.7 & 69.8 & 94.5  & 258.1
  & 17.1 & 40.6 & 53.7 & 85.5  & 196.9
  & 3.1  & 11.6 & 18.2 & 61.4  & 94.2 \\
ProtoPRVR~\cite{protoprvr}
  & 34.7 & 60.0 & 70.1 & 94.4  & 259.2
  & 16.0 & 38.8 & 52.4 & 85.1  & 192.3
  & -    & -    & -    & -     & -    \\
ARL~\cite{arl}
  & 34.6 & 60.4 & 70.7 & 94.4  & 260.1
  & 15.3 & 38.4 & 51.5 & 85.2  & 190.4
  & -    & -    & -    & -     & -    \\
\hlineB{2.5}\rowcolor{gray!20}
\textbf{Ours}
  & \textbf{35.1} & \textbf{61.6} & \textbf{71.5} & \textbf{94.9}  & \textbf{263.1}
  & \textbf{17.7} & \textbf{42.0} & \textbf{55.6} & \textbf{86.8}  & \textbf{202.1}
  & 3.2  & \textbf{12.6} & \textbf{20.1} & \textbf{63.8}  & \textbf{99.7} \\
\hlineB{2.5}
\end{tabular}
\vspace{-0.1cm}
\end{table*}
\endgroup

%% file: table/tab3_sentence_sim.tex
% \begin{table}[t]
%     \small
%     \centering
%     \caption{Similarity comparison between text(video) instances associated with each video.
%     Intra Sim represents the average similarity of instances paired with the same video while Total Sim is the average pair-wise similarity of all instances.
%     Diff.Norm is calculated as (Intra Sim - Total Sim) / (Intra Sim + Total Sim) to report the similarity gap between Intra Sim and Total Sim.
%     }
%     \label{tab:semantic_collapse_sim}
%     \renewcommand{\arraystretch}{1.0} % 행 간격 조절
%     \begin{tabular}{l|c|c|c|c|c|c|c|c|c}
%         \hlineB{2.5}
%          % & \multicolumn{2}{c}{\textbf{TVR}} \\  
%         % \cmidrule(lr){2-3} 
%         Method & \multirow{all}{Text로 채워 row 전부} & Intra Sim & Total Sim & Diff.Norm & \multirow{all}{Video로 채워 row 전부} & Intra Sim & Total Sim & Diff.Norm\\  
%         \hline
%         % CLIP & text & 0.6124 & 0.5633 & 0.0418 & video & - & - & - \\
%         GMM & text & 0.1175 & 0.0113 & 0.8245 & video & 0.6419 & 0.0623 & 0.8230 \\ 
%         GMMv2 & text & 0.1646 & 0.0196 & 0.7872 & video & 0.6041 & 0.0387 & 0.8796 \\
%         Ours & text & 0.2198 & 0.0813 & 0.4600 & video & 0.5531 & 0.0812 & 0.7440 \\
%         \hlineB{2.5}
%     \end{tabular}
% \end{table}
% \FloatBarrier
\begin{table}[h]
\vspace{-0.3cm}
    \footnotesize
    \centering
    \caption{Semantic similarity comparison between text and video instances per video. \textit{Intra Sim} is the average similarity among instances of the same video, \textit{Total Sim} is the average pairwise similarity across all instances, and \textit{Diff. Norm} is computed as \((\text{Intra Sim} - \text{Total Sim}) / (\text{Intra Sim} + \text{Total Sim})\) to represent the normalized gap between Intra Sim and Total Sim.}
    \label{tab:semantic_collapse_sim}
    \setlength{\tabcolsep}{3.1pt}
    \renewcommand{\arraystretch}{0.95}
    \begin{tabular}{l|cccc|cccc}
        \hlineB{2.5}
        % Method   & \multicolumn{3}{c|}{Text}                     & \multicolumn{3}{c}{Video}                    \\
        Method & \multicolumn{1}{c|}{Modality} & Intra Sim & Total Sim & Diff. Norm & \multicolumn{1}{c|}{Modality} & Intra Sim & Total Sim & Diff. Norm    \\
        \hline % \cline{1-1} \cline{3-5} \cline{7-9} 
        % \hline
        % CLIP~\cite{clip_ID} & \multicolumn{1}{c|}{~} & 0.6124    & 0.5633 & 0.0418 & \multicolumn{1}{c|}{~} & 0.7113    & 0.5356 & 0.1409     \\ 
        GMMF~\cite{gmm_PRVR} & \multicolumn{1}{c|}{\multirow{3}{*}{Text}} & 0.1175    & 0.0113    & 0.8245 & \multicolumn{1}{c|}{\multirow{3}{*}{Video}} & 0.6419    & 0.0623    & 0.8230        \\ 
        GMMF-v2~\cite{gmmv2_PRVR} & \multicolumn{1}{c|}{~} & 0.1646    & 0.0196    & 0.7872 & \multicolumn{1}{c|}{~} & 0.6041    & 0.0387    & 0.8796        \\ 
        % \cline{1-1} \cline{3-5} \cline{7-9}
        % CLIP~\cite{clip_ID} + OP-ToMe & \multicolumn{1}{c|}{~} & - & - & - & \multicolumn{1}{c|}{~} & 0.6696 & 0.5166 & 0.1290     \\ 
        Ours & \multicolumn{1}{c|}{~} & 0.2198    & 0.0813    & 0.4600 & \multicolumn{1}{c|}{~}    & 0.5531    & 0.0812    & 0.7440        \\
        \hlineB{2.5}
    \end{tabular}
    \vspace{-0.3cm}
\end{table}

%% file: table/tab4_inter_intra_text_sim.tex
\begin{wraptable}{r}{0.3\textwidth}
    \vspace{-0.4cm}
    % \caption{Spearman's rank correlation between PRVR models and CLIP.
    % Higher scores indicate a semantic structure more aligned with CLIP.
    % }
    \caption{Spearman's rank correlation with CLIP.
    % ; higher is more semantically aligned.
    }
    \label{tab:spearman}
    \vspace{-0.2cm}
    \centering
    \footnotesize
    \setlength{\tabcolsep}{10pt}
    \renewcommand{\arraystretch}{0.92}
    \begin{tabular}{l|c}
        \hlineB{2.5}
        Method & CLIP \\ \hline \hline
        Baseline & 35.40 \\
        MS-SL~\cite{ms-sl_PRVR} & 37.17 \\
        GMMF~\cite{gmm_PRVR} & 36.06 \\
        GMMF-v2~\cite{gmmv2_PRVR} & 35.74 \\ \hline
        Ours & \textbf{68.18} \\
        \hlineB{2.5}
    \end{tabular}
\vspace{-0.5cm}
\end{wraptable}

%% file: sec/5_conclusion.tex
\section{Conclusion \& Limitation}
% In this paper, we address the issue of overfitting in both text and video modalities: inter- and intra- semantic collapse in text representations and the attention collapse problem when encoding untrimmed videos.
% To tackle these challenges, we propose TCPL and SRT.
% Particularly, TCPL mitigates the semantic collapse by transferring the relational knowledge from the foundation model.
% In addition, SRT alleviates attention collapse, where video tokens become overly dependent on specific frames, by adjusting the attention map based on temporal correlations produced within the same embedding space.
% Through various quantitative and qualitative evaluations, we validate the effectiveness of our approach and demonstrate its improved performance over existing PRVR models.
% These findings underscore the importance of addressing semantic and attention collapse in the PRVR task.
% TCPL and SRT mitigate semantic and attention collapse by leveraging well-preserved text relational knowledge and self-relational knowledge, respectively.
% Through qualitative and quantitative evaluations, we validate the effectiveness of our approach and demonstrate its improved performance over existing PRVR models.
% These findings underscore the importance of addressing semantic and attention collapse in the PRVR task.
\textbf{Conclusion.} In this paper, we address semantic collapse in PRVR, where semantically diverse text queries and video segments are undesirably attracted or repelled due to pairwise annotation schemes. 
To mitigate this, we propose a unified framework consisting of Text Correlation Preservation Learning~(TCPL) and Cross-Branch Video Alignment~(CBVA). 
TCPL distills the relational structure from CLIP to preserve semantic consistency across text queries, while CBVA aims to structure video embeddings according to their inherent semantics, supported by our token merging strategies.
Extensive evaluations highlight the importance of addressing semantic collapse for effective PRVR.
% Through extensive evaluations, we validate the effectiveness of our approach.
% These findings underscore the importance of addressing both the text and video semantic collapse in PRVR.

\textbf{Limitation.}
% Our framework builds directly on CLIP’s pretrained text–image alignment.
% Therefore, it is challenging to retrieve segments involving subtle visual cues or directional queries.
% However, in the appendix, we report that CLIP의 failure case 중 multi-entity context와 multi event temporal composition 등과 관련해서는 많은 부분 failure case를 해결해서 총 R1 기준 28\% 해결하고, R10 기준 57\% 해결한다.
% 또한, 우리는 branch 간의 fine-grained alignment 학습 과정이 추가되기때문에 학습 코스트가 추가된다.
% 하지만, 실제 deploymen와 관련된 inference time과 관련해서는 증가가 없다.
Our method has two limitations.
First, as our method builds upon the pretrained CLIP model, it can inherit weaknesses; it may struggle with fine-grained spatial/directional queries~(e.g., distinguishing "left of" from "right of"). 
However, we emphasize that this limitation does not extend to compositional understanding. 
As we demonstrate in the Appendix, our method actively corrects CLIP's common failure modes where the queries involve multi-entity contexts and multi-event temporal compositions~(recovering 28\% of CLIP's $R@1$ failure cases and 57\% of its $R@10$ failure cases).
Second, our framework incurs an increased training cost. 
% This trade-off, however, is confined entirely to the training phase. 
However, for deployment, our model architecture does not introduce any new modules that increase inference time, incurring no additional latency compared to standard retrieval baselines.

%% file: main_appendix.tex
\renewcommand{\thesection}{A}   
\renewcommand{\thetable}{A\arabic{table}}   
\renewcommand{\thefigure}{A\arabic{figure}}
\setcounter{section}{0}
\setcounter{page}{1}
\setcounter{table}{0}
\setcounter{figure}{0}

% \section{Datasets}
% As described in the main manuscript, we evaluate our method on four untrimmed text-video datasets. 
% Below, we provide a brief overview of each.
% QVHighlights\cite{qvhighlight} is a collection of news and vlog-style videos, recently reorganized for PRVR\cite{protoprvr}. 
% Each video is paired with an average of 3.3 text queries describing semantically diverse segments.
% TVR~\cite{tvr_DATA} is built from scenes across six popular TV shows, with each video annotated by five text queries targeting different segments. 
% The training set contains 17,435 videos and 87,175 queries, while the evaluation set includes 2,179 videos and 10,895 queries.
% ActivityNet Captions~\cite{anetcaptions_DATA} is sourced from YouTube videos, with an average of 3.7 text queries per video. 
% The dataset includes 10,009 videos for training and 4,917 for evaluation.
% Charades-STA~\cite{charades_DATA} extends the original Charades dataset by adding sentence-level annotations for specific temporal segments. 
% It consists of 13,898 video-sentence pairs for training and 4,233 for evaluation.

\begin{table}[h!]
\centering
\caption{Sensitivity to temperature $\tau$ across datasets. Rows marked with gray indicate the default configuration used in the main results.}
\label{tab:tau_sensitivity}
\setlength{\tabcolsep}{5pt}
\renewcommand{\arraystretch}{0.95}
\begin{tabular}{lcccccc}
\toprule
Dataset & $\tau$ & R@1 & R@5 & R@10 & R@100 & SumR \\
\midrule
\multirow{5}{*}{TVR}
 & 0.70 & 35.6 & 61.0 & 70.8 & 95.0 & 262.4 \\
 & 0.75 & 35.5 & 61.2 & 71.1 & 94.9 & 262.6 \\ \rowcolor{gray!20}
 TVR & 0.80 & 35.1 & 61.6 & 71.5 & 94.9 & 263.1 \\
 & 0.85 & 35.1 & 61.2 & 71.2 & 95.0 & 262.5 \\
 & 0.90 & 35.1 & 61.1 & 71.1 & 94.9 & 262.2 \\
\midrule
\multirow{5}{*}{ANet}
 & 0.70 & 17.6 & 41.9 & 55.4 & 86.8 & 201.7 \\
 & 0.75 & 17.8 & 41.9 & 55.4 & 86.7 & 201.8 \\ \rowcolor{gray!20}
 ANet & 0.80 & 17.7 & 42.0 & 55.6 & 86.8 & 202.1 \\
 & 0.85 & 17.7 & 42.1 & 55.3 & 86.8 & 201.9 \\
 & 0.90 & 17.2 & 41.9 & 55.5 & 86.8 & 201.4 \\
\midrule
\multirow{5}{*}{CHA}
 & 0.70 & 3.3 & 11.6 & 19.8 & 63.9 & 98.6 \\
 & 0.75 & 3.4 & 12.7 & 19.4 & 64.8 & 100.3 \\
 & 0.80 & 3.4 & 12.0 & 18.7 & 64.5 & 98.6 \\
 \rowcolor{gray!20}
 & 0.85 & 3.2 & 12.6 & 20.1 & 63.8 & 99.7 \\
 & 0.90 & 3.3 & 12.4 & 19.1 & 64.0 & 98.9 \\
\bottomrule
\end{tabular}
\end{table}
\renewcommand{\thesection}{A}   
\section{Further Analysis on Hyperparameter Sensitivity}
\label{sec:hparam_sensitivity}
We noted that all hyperparameters are unified across datasets except the similarity threshold $\tau$, which we set per dataset to account for different internal segment-to-segment similarity distributions~\cite{protoprvr}.
Beyond the QVHighlights ablation, Table~\ref{tab:tau_sensitivity} evaluates $\tau$ sensitivity on TVR, ActivityNet-Captions~(ANet), and Charades as well.
Empirically, QVHighlights exhibits the lowest similarity levels, TVR and ANet are intermediate, and CHA shows the highest.
Accordingly, we adopt $\tau{=}0.70$ for QVHighlights, $\tau{=}0.80$ for TVR and ANet, and $\tau{=}0.85$ for CHA.
As shown, varying $\tau$ within a moderate range causes only minor fluctuations in each dataset, indicating that performance is not overly sensitive to this hyperparameter once set near the optimum.

\begin{table}[t]
\small
\centering
\caption{
Comparative analysis of retrieval correctness between our model and zero-shot CLIP on the TVR test set (10,895 queries), evaluated using (a) Recall@1 and (b) Recall@10 as success criteria. Values are raw counts with percentages in parentheses.
}
\label{tab:tvr_clip_confusion}
\setlength{\tabcolsep}{7pt}
\renewcommand{\arraystretch}{1.05}
\begin{subtable}[t]{0.46\textwidth}
\centering
\caption{Recall@1.}
\begin{tabular}{lcc}
\toprule
 & CLIP correct & CLIP wrong \\
\midrule
Ours correct & 1277 (11.7\%) & 2551 (23.4\%) \\
Ours wrong   & 500 (4.6\%)  & 6567 (60.3\%) \\
\bottomrule
\end{tabular}
\end{subtable}\hfill
\begin{subtable}[t]{0.46\textwidth}
\centering
\caption{Recall@10.}
\begin{tabular}{lcc}
\toprule
 & CLIP correct & CLIP wrong \\
\midrule
Ours correct & 4162 (38.2\%) & 3627 (33.3\%) \\
Ours wrong   & 386 (3.5\%)  & 2720 (24.9\%) \\
\bottomrule
\end{tabular}
\end{subtable}
\end{table}

%%% ### --- CLIP Impact on TCPL ---- ### %%%
\renewcommand{\thesection}{B}
\section{Impact of CLIP's Failure Rate on TCPL}
\label{sec:clip_failure_tcpl}
In this section, we evaluate whether TCPL inherits or corrects CLIP’s semantic errors in the PRVR setting.
We conduct this study on the TVR dataset since most text queries in TVR involve multiple named entities or sequential actions that require the capability to comprehend complex temporal and contextual cues. 
On the test set of TVR~(10,895 queries), we mark a success when the ground-truth video appears within the top-$Q$ retrieved results~($Q\in\{1,10\}$) and compare our model~(with TCPL) to zero-shot CLIP via a $2{\times}2$ outcome matrix.
Specifically, for each text query, we record (i) both correct, (ii) ours correct \& CLIP wrong, (iii) ours wrong \& CLIP correct, and (iv) both wrong.
Tab.~\ref{tab:tvr_clip_confusion} reports the counts~(and proportions).

To illustrate, when $Q{=}1$, our model corrects 2,551 of CLIP’s failures~(while the reverse occurs in 500 cases); at $Q{=}10$, the corresponding counts are 3,627 vs. 386.
Our proposed framework also retains CLIP’s strengths, answering correctly together on 1,277~(R@1) and 4,162~(R@10) queries.

We further analyze the instances where one model succeeds and the other fails.
When CLIP fails, the correct item is, on average, ranked 56th, indicating severe confusion. 
These failures consistently involve queries with multi-entity contexts and temporal compositions. 
For example, CLIP ranked the correct video at 237 for “Sebastian grabs his folder and stands up from the table” and at 418 for “George pulls back on Meredith’s rolling chair and drags her”. 
By contrast, when our model fails but CLIP succeeds, the ground-truth video is still ranked highly, with an average position of 6.7. 
These cases are typically simple and object-centric queries requiring little compositional or temporal reasoning. 
For instance, CLIP correctly retrieved the videos for “House takes a sip of soda from the bottle” and “Joey is folding his coat in the kitchen”, while our model placed them at rank 2. 
Taken together, these outcomes demonstrate that the retrieval objective reshapes the representation toward task-specific temporal and compositional semantics, with TCPL preserving robust high-level alignment while correcting CLIP’s fine-grained failure modes.

\input{algorithm/alg}

\renewcommand{\thesection}{C}
\section{Algorithms for Cross-Branch Video Alignment}
In this section, we provide a detailed algorithm for sub-components of our Cross-Branch Video Alignment~(CBVA).
Particularly, we illustrate Order-Preserving Token Merging~(OP-ToMe), the process of pre-computing a discrete set of different levels of clip number~(number of semantics), and the process of per-video merging for Adaptive CBVA in Algorithm.~\ref{alg.optome}, Algorithm.~\ref{alg.cliplevel}, and Algorithm.~\ref{alg.adaptiveCBVA}, respectively.

\input{algorithm/alg2}

\renewcommand{\thesection}{D}   
\section{Positive and Negative Societal Impacts}

\textbf{Positive Impact.}  
Our work improves the text-video retrieval based on partial content descriptions within long, untrimmed videos.
We expect that the proposed method will enhance the user experience in video search and navigation. 
This is particularly valuable in domains such as education, where lengthy untrimmed videos are commonly utilized.

\textbf{Negative Impact.}  
However, the ability to isolate specific video contexts and retrieve segments based on partial descriptions could be misused in surveillance settings~(e.g., CCTV), enabling the tracking of individuals or the extraction of sensitive behaviors without consent. 
Such misuse may raise potential concerns regarding privacy and ethical deployment.

\newpage

%% file: algorithm/alg.tex
% OP-TOME algorithm
\begin{algorithm}[t]
\caption{Order-Preserving Token Merging (OP-ToMe)}
\label{alg.optome}
\begin{algorithmic}[1]
\Require Frame tokens $V_f \in \mathbb{R}^{B_v \times L_f \times d_v}$, Merge rate $N\%$, Number of iterations $M$
\Ensure Clip tokens $V_c \in \mathbb{R}^{B_v \times L_c \times d_v}$ where $L_c = 32$
\State Initialize token sizes $s \leftarrow \textbf{1}_{L_f} \in \mathbb{R}^{L_f}$  \Comment{Each token represents 1 frame}
\For{$m = 1$ to $M$}
    \State Compute cosine similarity between disjoint adjacent-frame pairs:
    \Statex \hspace{\algorithmicindent} $S[i] \leftarrow \cos(V_f[i], V_f[i+1])$ for $i = 1, 3, 5, \ldots, L_f - 1$
    \State Select top-$N\%$ most similar adjacent pairs based on $S$
    \For{each selected pair $(i, i+1)$}
        \State Compute size-weighted average: 
        \Statex \hspace{\algorithmicindent} $V_{\text{merged}} \leftarrow \frac{s[i] \cdot V_f[i] + s[i+1] \cdot V_f[i+1]}{s[i] + s[i+1]}$
        \State Replace $V_f[i]$ with $V_{\text{merged}}$, remove $V_f[i+1]$
        \State Update size: $s[i] \leftarrow s[i] + s[i+1]$, remove $s[i+1]$
    \EndFor
    \State Update $L_f \leftarrow$ new token length
    \If{$L_f \leq 32$}
        \State \textbf{break}
    \EndIf
\EndFor
\State \Return $V_c \leftarrow V_f$
\end{algorithmic}
\end{algorithm}
\vspace{-0.3cm}
\begin{algorithm}[t]
\caption{Pre-computing the different levels of clip number~(Eq.~\ref{eq.cliplevel})}
\label{alg.cliplevel}
\begin{algorithmic}[1]
\Require Initial clip length $L_c^{1}=L_c$ (e.g., 32), merge‑rate $N\%$, minimum clips $C_{\min}$
\Ensure  Candidate list $L = \bigl[L_c^{1},L_c^{2},\dots,L_c^{K}\bigr]$
\State $i \leftarrow 1$, \; $L\,\leftarrow\,[\,L_c^{1}\,]$
\While{$L_c^{i} > C_{\min}$}
    \State $\displaystyle
           L_c^{\,i+1} \leftarrow
           \max\!\Bigl(
               2 \times
               \Bigl\lfloor
                   \frac{\,L_c^{i}- (L_c^{i}/2)\,(N/100) + 1\,}{2}
               \Bigr\rfloor,\;
               C_{\min}
           \Bigr)$
    \If{$L_c^{\,i+1} = L_c^{i}$} \textbf{break} \EndIf
    \State Append $L_c^{\,i+1}$ to ${L}$
    \State $i \leftarrow i+1$
\EndWhile
\State $K \leftarrow |{L}|$ \Comment{number of discrete clip levels}
\State \Return ${L}$
\end{algorithmic}
\end{algorithm}
% ---------- Per‑video procedure ----------

%% file: algorithm/alg2.tex
\begin{algorithm}[t]
\caption{Constructing merged clips for Adaptive CBVA}
\label{alg.adaptiveCBVA}
\begin{algorithmic}[1]
\Require Clip tokens $V_c \!\in\! \mathbb{R}^{B_v\times L_c \times d_v}$, 
         Global candidate list $L$ of length $K$,  
         Merge rate $N\%$, Similarity threshold $\tau$,
         Projected Clip tokens $\bar{V}_c \!\in\! \mathbb{R}^{B_v\times L_c \times d}$,  
\Ensure  Adapted clip tokens $\tilde{V}_c$ with length $L_c^\ast$

\noindent\textbf{Stage 1. Estimate internal similarity}
\State Compute cosine‑similarity matrix $S$ from \emph{frozen} $V_c$
\State $\displaystyle
       \omega \gets
       \frac{\bigl|\{(i,j)\!:\!S_{ij}\!>\!\tau,\; i\!\neq\! j\}\bigr|}
            {L_c(L_c-1)}
      $ \Comment{high‑similarity ratio}

\noindent\textbf{Stage 2. Select merging depth $k^{\ast}$}
\If{$\omega \le 1-\tfrac{1}{K}$} \Comment{if diverse, keep all clips}
    \State $k^{\ast} \gets 1$
\Else
    \State $k^{\ast} \gets 
        % \min\bigl\{\,k\in\{2,\dots,K\}\;\big|\;
        %            \omega > \tfrac{K-k}{K}\bigr\}$
        \min_{k \in \{2, \cdots, K\}} (w > \frac{K-k}{K}) $
\EndIf

\noindent\textbf{Stage 3. Merge clips $k^\ast\!-\!1$ times}
\State $\tilde{V}_c \gets \bar{V}_c$
\For{$m = 1$ \textbf{to} $k^\ast-1$}
    \State Apply \emph{bipartite token merging (TOME)}~\cite{tome} to $\tilde{V}_c$ at rate $N\%$
\EndFor
\State $L_c^\ast \gets |\tilde{V}_c|$
\State \Return $\tilde{V}_c$
\end{algorithmic}
\end{algorithm}